\PassOptionsToPackage{table,dvipsnames}{xcolor}
\documentclass[11pt, a4paper, logo, copyright]{eai}
\usepackage{shlab}

\usepackage[authoryear, sort&compress, round]{natbib}
\usepackage{amsmath,amsfonts,bm}

\makeatletter
\newcommand{\defmacro}[1]{\let#1\@undefined\newcommand{#1}}
\makeatother

\defmacro{\figleft}{{\em (Left)}}
\defmacro{\figcenter}{{\em (Center)}}
\defmacro{\figright}{{\em (Right)}}
\defmacro{\figtop}{{\em (Top)}}
\defmacro{\figbottom}{{\em (Bottom)}}
\defmacro{\captiona}{{\em (a)}}
\defmacro{\captionb}{{\em (b)}}
\defmacro{\captionc}{{\em (c)}}
\defmacro{\captiond}{{\em (d)}}

\defmacro{\newterm}[1]{{\bf #1}}

\def\eqref#1{\cref{#1}}

\def\1{\bm{1}}
\defmacro{\train}{\mathcal{D}}
\defmacro{\valid}{\mathcal{D_{\mathrm{valid}}}}
\defmacro{\test}{\mathcal{D_{\mathrm{test}}}}

\DeclareMathAlphabet{\mathsfit}{\encodingdefault}{\sfdefault}{m}{sl}
\SetMathAlphabet{\mathsfit}{bold}{\encodingdefault}{\sfdefault}{bx}{n}
\defmacro{\tens}[1]{\bm{\mathsfit{#1}}}

\defmacro{\etens}[1]{\mathsfit{#1}}

\defmacro{\pdata}{p_{\rm{data}}}
\defmacro{\ptrain}{\hat{p}_{\rm{data}}}
\defmacro{\Ptrain}{\hat{P}_{\rm{data}}}
\defmacro{\pmodel}{p_{\rm{model}}}
\defmacro{\Pmodel}{P_{\rm{model}}}
\defmacro{\ptildemodel}{\tilde{p}_{\rm{model}}}
\defmacro{\pencode}{p_{\rm{encoder}}}
\defmacro{\pdecode}{p_{\rm{decoder}}}
\defmacro{\precons}{p_{\rm{reconstruct}}}

\defmacro{\laplace}{\mathrm{Laplace}} 

\defmacro{\E}{\mathbb{E}}
\defmacro{\Ls}{\mathcal{L}}
\defmacro{\R}{\mathbb{R}}
\defmacro{\emp}{\tilde{p}}
\defmacro{\lr}{\alpha}
\defmacro{\reg}{\lambda}
\defmacro{\rect}{\mathrm{rectifier}}
\defmacro{\softmax}{\mathrm{softmax}}
\defmacro{\sigmoid}{\sigma}
\defmacro{\softplus}{\zeta}
\defmacro{\KL}{D_{\mathrm{KL}}}
\defmacro{\Var}{\mathrm{Var}}
\defmacro{\standarderror}{\mathrm{SE}}
\defmacro{\Cov}{\mathrm{Cov}}
\defmacro{\normlzero}{L^0}
\defmacro{\normlone}{L^1}
\defmacro{\normltwo}{L^2}
\defmacro{\normlp}{L^p}
\defmacro{\normmax}{L^\infty}

\defmacro{\parents}{Pa} 

\usepackage{multirow}
\usepackage{makecell}
\usepackage{array}
\usepackage{subcaption}
\usepackage{tikz}
\usetikzlibrary{backgrounds}
\usepackage{pifont}
\usepackage{xspace}

\makeatletter
\let\ego@standardaddtocurcol\@addtocurcol
\makeatother
\usepackage{flafter}
\usepackage{float} 
\usepackage{placeins}
\usepackage{afterpage}

\renewcommand{\topfraction}{0.92}
\renewcommand{\textfraction}{0.07}
\renewcommand{\floatpagefraction}{0.75}

\hypersetup{colorlinks=true,linkcolor=NavyBlue,citecolor=NavyBlue,urlcolor=NavyBlue}
\usepackage[nameinlink]{cleveref}

\graphicspath{{figures/}}
\setkeys{Gin}{width=0.99\linewidth}
\crefname{figure}{Fig.}{Figs.}
\Crefname{figure}{Fig.}{Figs.}
\crefname{subfigure}{Fig.}{Figs.}
\Crefname{subfigure}{Fig.}{Figs.}
\crefname{table}{Tab.}{Tabs.}
\Crefname{table}{Tab.}{Tabs.}
\crefname{section}{Sec.}{Secs.}
\Crefname{section}{Sec.}{Secs.}
\crefname{subsection}{Sec.}{Secs.}
\Crefname{subsection}{Sec.}{Secs.}
\crefname{subsubsection}{Sec.}{Secs.}
\Crefname{subsubsection}{Sec.}{Secs.}
\crefname{appendix}{App.}{Apps.}
\Crefname{appendix}{App.}{Apps.}
\crefname{equation}{Eq.}{Eqs.}
\Crefname{equation}{Eq.}{Eqs.}

\newif\ifdraftmode
\draftmodefalse         

\ifdraftmode
  \defmacro{\todo}[1]{{\color{red}\small\sffamily [TODO: #1]}}
  \defmacro{\note}[2]{{\color{blue}\small\sffamily [#1: #2]}}
\else
  \defmacro{\todo}[1]{}
  \defmacro{\note}[2]{}
\fi

\ifdraftmode
  \defmacro{\PH}[1]{{\color{orange!85!black}\textbf{#1}}}
\else
  \defmacro{\PH}[1]{#1}
\fi
\defmacro{\pmch}{\text{--}}          
\defmacro{\na}{\textcolor{gray}{--}} 

\newlength{\phwd}\newlength{\phht}\newlength{\phtw}
\defmacro{\figph}[3]{%
  \setlength{\phwd}{#1}%
  \setlength{\phht}{#2}%
  \setlength{\phtw}{\phwd}\addtolength{\phtw}{-18pt}%
  \begin{tikzpicture}
    \node[draw=gray!65,dashed,line width=0.6pt,fill=gray!8,rounded corners=3pt,
          minimum width=\phwd,minimum height=\phht,
          text width=\phtw,align=center,inner sep=6pt,
          font=\footnotesize\sffamily\color{gray!35!black}] {#3};
  \end{tikzpicture}%
}

\defmacro{\name}{\textsc{EgoExplorer}\xspace}
\defmacro{\smplx}{SMPL-X\xspace}
\defmacro{\mano}{MANO\xspace}
\defmacro{\srv}{SRV\xspace}                 
\defmacro{\locoman}{loco-manipulation\xspace}
\defmacro{\evalsubset}{Evaluation Subset\xspace}
\defmacro{\naturaldata}{Natural Recordings\xspace}
\defmacro{\evalshort}{Evaluation\xspace}
\defmacro{\naturalshort}{Natural\xspace}
\defmacro{\dof}{DoF\xspace}

\defmacro{\mm}{\,mm}
\defmacro{\cm}{\,cm}
\defmacro{\ms}{\,ms}
\defmacro{\hz}{\,Hz}
\defmacro{\degr}{$^{\circ}$}
\defmacro{\hours}{\,h}

\defmacro{\best}[1]{\textbf{#1}}
\defmacro{\second}[1]{\underline{#1}}
\defmacro{\ours}{\rowcolor{blue!6}}
\newlength{\summarycolwidth}
\defmacro{\up}{$\uparrow$}
\defmacro{\dn}{$\downarrow$}
\defmacro{\cmark}{\ding{51}}
\defmacro{\xmark}{\ding{55}}
\defmacro{\pmark}{$\circ$}   

\title{\name: A Vision-Motion-Touch Observatory for Human Loco-Manipulation In Situ}

\author[$*$,$\dagger$,1]{Yunsong Zhou}
\author[$*$,1]{Xuekun Jiang}
\author[$*$,1]{Haiqi Shen}
\author[1]{Hangxu Liu}
\author[1]{\\Chaojian Shi}
\author[1]{Yujie Liu}
\author[1]{Jingyuan Li}
\author[1]{Haoran Yang}
\author[1]{Jiafei Cao}
\author[1]{\\Dongzhan Zhou}
\author[1]{Dong Wang}
\author[1]{Yao Mu}
\author[1,2]{Chunhua Shen}
\author[$\dagger$,1,3]{Weinan Zhang}

\affil[1]{Shanghai AI Laboratory}
\affil[2]{Zhejiang University}
\affil[3]{Shanghai Jiao Tong University}

\correspondingauthor{$\dagger$\,Corresponding authors.}

\def\urlhomepage{https://internrobotics.github.io/egoexplorer.github.io/}
\def\urltoolkit{https://github.com/InternRobotics/EgoExplorer}
\def\urldata{https://huggingface.co/datasets/InternRobotics/EgoExplorer}

\renewcommand{\absfont}{\linespread{1.08}\itshape\fontsize{10}{12}\selectfont}
\makeatletter
\patchcmd{\maketitle}{\vskip15pt}{\vskip4pt}{}%
  {\PackageWarning{egoexplorer}{maketitle skip not patched}}
\makeatother

\begin{document}

\begin{abstract}
Learning humanoid \locoman from human demonstrations rests on observing how locomotion, posture, and object interaction evolve together as tasks unfold.
Scaling collection across diverse environments, in turn, favors lightweight, portable capture systems.
Yet, existing datasets provide only partial views of this coordination.
Thus, we introduce \name, a 100-hour vision-motion-touch observatory of human \locoman built with a wearable system.
With an emphasis on professional activities, its continuous recordings align egocentric video with reconstructed whole-body motion, fine-grained finger articulation, and tactile observations, accompanied by step-level descriptions.
To achieve this coverage with sparse body sensing, we develop the \emph{Calibrate-Complete-Constrain} pipeline.
Concretely, reusable calibration fixes participant geometry and sensor attachments before motion inference to disentangle mounting offsets from articulation. 
A temporal prior then completes unobserved joint motion, while layered constraints promote measurement agreement, temporal continuity, and geometric consistency, properties that motion plausibility alone does not guarantee.
Against tracker-fitted references, the method reduces body-point error by 71.9\% relative to the strongest baseline; matched evaluations also show gains in hand stability and contact estimation.
Preliminary studies explore multimodal generation and conditioning alongside humanoid retargeting with whole-body control in simulation.
\end{abstract}

\maketitle
\vspace{-0.15cm}

\links{%
  \link{homepage}{Homepage}{\urlhomepage},
  \link{code}{Toolkit}{\urltoolkit},
  \link{data}{Data}{\urldata}%
}

\vspace{-0.35cm}
\noindent\begin{minipage}{\textwidth}
  \centering
  \includegraphics[width=0.96\textwidth]{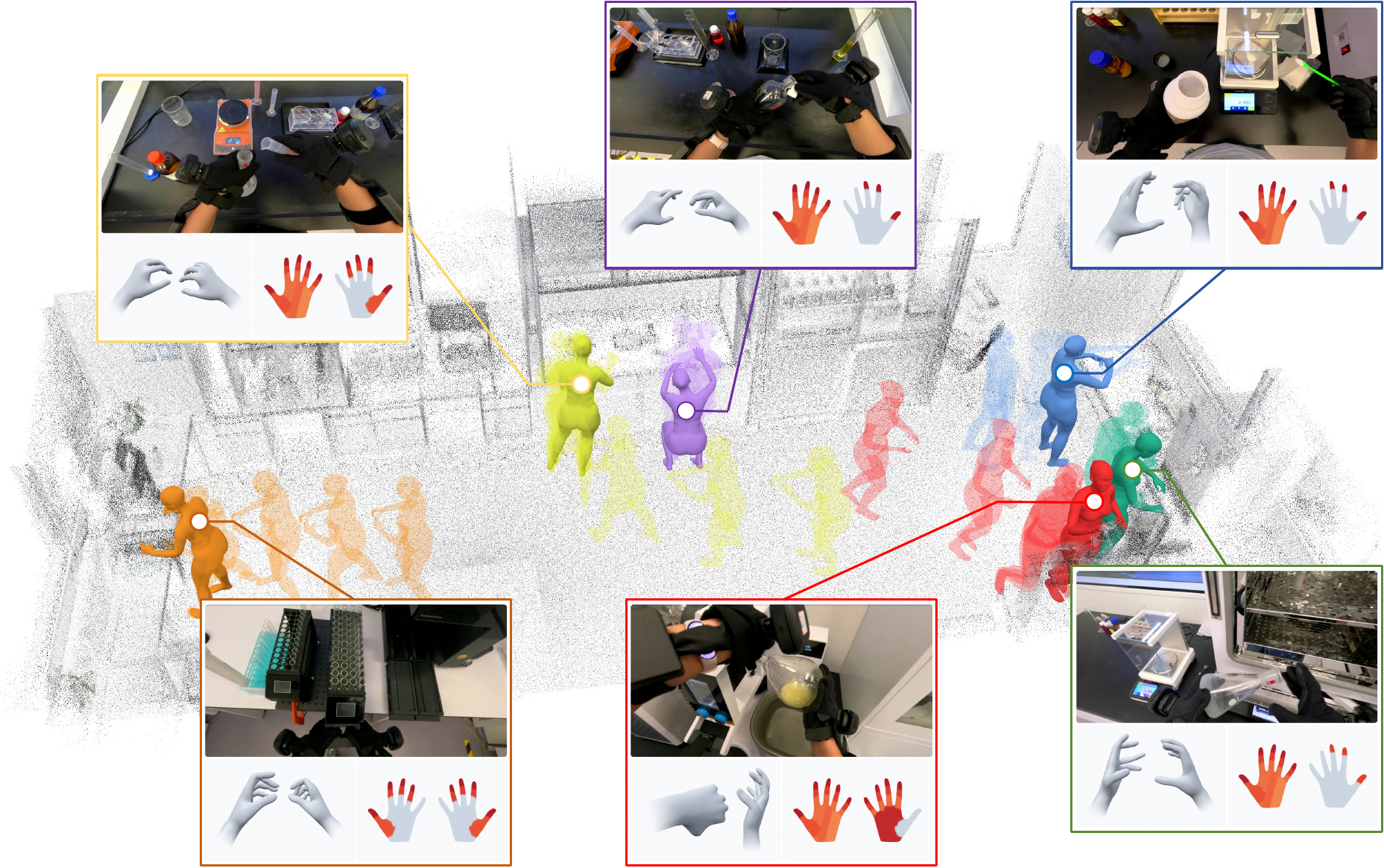}
  \captionof{figure}{
  \textbf{A glimpse of \name}. It shows human loco-manipulation across workstations, with reconstructed body motion overlaid on the scene point cloud. Each inset is linked to a corresponding body pose along with synchronized egocentric view, finger articulation, and tactile observation.}
  \label{fig:teaser}
\end{minipage}
\par\medskip

\vspace{-0.2cm}

\section{Introduction}
\label{sec:intro}

What a humanoid needs to reach, grasp, or carry shapes where it moves and how it positions its body \citep{shi2026egohumanoid}.
Learning such \locoman skills calls for demonstrations spanning diverse environments, objects, and task sequences.
Robot teleoperation provides such
data, but dedicated hardware, on-site deployment, and sustained operator
effort constrain its scalability \citep{zhu2025emma,punamiya2026egoverse}.
Human activity offers a rich alternative across everyday and professional
settings, and recent work shows that egocentric demonstrations can improve
robot manipulation and generalization to novel scenarios
\citep{kareer2024egomimic,qiu2025ph2d,shi2026egohumanoid}.
Realizing this potential for \locoman hinges on datasets that jointly record visual context, whole-body and hand motion, and measured contact throughout continuous tasks.

Building such datasets at scale entails balancing environmental coverage with the richness and fidelity of the observations.
Systems built around fixed camera arrays for optical motion
capture \citep{cao2026acedata0} or multi-view reconstruction \citep{bhatnagar2022behave,zhang2022egobody} can provide detailed observations, but
require each recording space to be instrumented and calibrated
\citep{taheri2020grab,kim2024parahome}. Portable capture eases deployment,
yet existing datasets often offer only partial views of this coordination.
Particularly, egocentric video preserves diverse visual context but leaves much of the
body unobserved \citep{grauman2022ego4d}.
Wearable body tracking adds body motion, often without fine finger
articulation or physical contact measurements
\citep{ma2024nymeria,detone2026nymeriaplus}.
Conversely, hand-focused datasets capture detailed hand motion and, in some cases
touch, but typically omit the accompanying whole-body movement
\citep{hoque2025egodex,zhou2026touchanything,song2025opentouch}.
These partial records obscure the coupling central to \locoman and therefore fall short of capturing its execution as a coherent whole.


To this end, we introduce \name, an egocentric dataset that captures the whole-body
coordination underlying human \locoman through portable
vision-motion-touch sensing. 
As depicted in \cref{fig:teaser}, it aligns first-person video with
reconstructed body motion, glove-measured finger articulation, and tactile
observations. Collected across five environments with an emphasis on professional activities, the recordings preserve the transitions between approaching targets, adjusting posture, manipulating objects, and relocating.
Step-level descriptions link these synchronized observations to the actions and progression of each task.

To support this multimodal capture, our lightweight wearable rig combines five trackers, instrumented gloves, and a head-mounted camera.
Recovering whole-body trajectories requires inferring unobserved joint motion from these sparse device poses.
Yet direct fitting is ambiguous: each pose depends jointly on its attachment transform and the body's kinematic chain, allowing different combinations of mounting offsets and joint configurations to explain the same sparse measurements \citep{jiang2022avatarposer,ponton2023sparseposer}.
%
Differently, our $\mathrm{C}^3$ pipeline (\emph{Calibrate-Complete-Constrain}) disentangles fixed body-sensor geometry from motion through a calibration protocol: prescribed participant movements are matched to a reference motion library.
Using these correspondences, we refine initial estimates of body shape and scale and fit body-sensor transforms, forming a profile reusable across tasks with unchanged sensor placement.
A temporal prior model then completes unobserved articulation from the calibrated anchors, yielding an initial whole-body trajectory, \textit{i.e.}, the motion proposal.
We refine this proposal under layered constraints to promote measurement agreement and temporal continuity while penalizing self-interpenetration and ground penetration.
Finally, we incorporate hand motion, touch, and expert-guided step annotations and check multimodal consistency.
The resulting dataset comprises 2{,}715 egocentric recordings totaling 100 hours (\cref{tab:comparison}).

\begin{table}[!t]
\centering
\vspace{-15pt}
\caption{\textbf{Comparison of representative egocentric datasets.} In the
body and hand groups, \cmark{} is sensor-grounded label, \pmark{} a state
inferred from images or covering a subset, \xmark{} absence; elsewhere
\cmark/\pmark{} are full/partial support. \texttt{Anchors}: non-visual
measurements (wrists, feet, and fingers) unaffected by occlusion. \texttt{Mobile capture}: the wearer's root
travels rather than staying at one workstation. \texttt{Loco-manip.}: both are
measurable in the motion data. \texttt{Step labels}: sub-task annotations,
\cmark{} if written or checked by experts, \pmark{} if left as model output.
$^\dagger$: judged from the released sample.}
\label{tab:comparison}
\setlength{\tabcolsep}{3pt}
\scriptsize
\begin{tabular}{@{}lrccccccccc@{}}
\toprule
\multirow{3}{*}[-2pt]{Dataset} & \multirow{3}{*}[-2pt]{\makecell{Duration\\(h)}} &
\multicolumn{3}{c}{Body state} & \multicolumn{2}{c}{Hand state} &
\multirow{3}{*}[-2pt]{Touch} & \multirow{3}{*}[-2pt]{\makecell{Mobile\\capture}} &
\multirow{3}{*}[-2pt]{\makecell{Loco-\\manip.}} &
\multirow{3}{*}[-2pt]{\makecell{Step\\labels}} \\
\cmidrule(lr){3-5}\cmidrule(lr){6-7}
 & & \makecell{metric\\full body} & \makecell{non-visual\\anchors} &
\makecell{feet\\observed} & articulated & \makecell{non-visual\\anchors} &
& & & \\
\midrule
Ego4D \citep{grauman2022ego4d}          & 3{,}670 & \xmark & \xmark & \xmark & \xmark & \xmark & \xmark & \cmark & \xmark & \cmark \\
Ego-Exo4D \citep{grauman2024egoexo4d}   & 221 & \pmark & \xmark & \xmark & \pmark & \xmark & \xmark & \cmark & \cmark & \cmark \\
Nymeria \citep{ma2024nymeria}           & 300 & \cmark & \cmark & \cmark & \xmark & \xmark & \xmark & \cmark & \cmark & \cmark \\
EgoDex \citep{hoque2025egodex}          & 829 & \pmark & \xmark & \xmark & \pmark & \xmark & \xmark & \xmark & \xmark & \xmark \\
EgoVerse \citep{punamiya2026egoverse}   & 1{,}362 & \xmark & \xmark & \xmark & \pmark & \xmark & \xmark & \cmark & \xmark & \cmark \\
Open-AoE \citep{li2026openaoe}          & 2{,}000 & \xmark & \xmark & \xmark & \pmark & \xmark & \xmark & \xmark & \xmark & \cmark \\
EgoLive \citep{li2026egolive}           & 1{,}680 & \xmark & \xmark & \xmark & \pmark & \xmark & \xmark & \xmark$^\dagger$ & \xmark & \pmark \\
Xperience-10M \citep{ropedia2026xperience} & 1{,}059 & \pmark & \xmark & \xmark & \pmark & \xmark & \xmark & \xmark$^\dagger$ & \cmark & \pmark \\
Egocentric-10K \citep{buildai2025egocentric10k} & 10{,}000 & \xmark & \xmark & \xmark & \xmark & \xmark & \xmark & \xmark$^\dagger$ & \xmark & \xmark \\
EgoTouch \citep{zhou2026touchanything}  & 20 & \xmark & \pmark & \xmark & \cmark & \cmark & \cmark & \xmark & \xmark & \xmark \\
OpenTouch \citep{song2025opentouch}     & 5 & \xmark & \xmark & \xmark & \cmark & \cmark & \cmark & \xmark & \xmark & \cmark \\
\midrule
\ours \name                          & 100 & \cmark & \cmark & \cmark & \cmark & \cmark & \cmark & \cmark & \cmark & \cmark \\
\bottomrule
\end{tabular}
\vspace{-15pt}
\end{table}

We evaluate annotation accuracy on a purpose-built \evalsubset and
reconstruction consistency on the \naturaldata. Compared with the strongest
evaluated body-reconstruction baseline, our system reduces body-point error
by 71.9\%. Hand and contact tests also show better holding stability,
kinematic target recovery, and regional contact estimation than the evaluated
alternatives, supporting the benefits of calibrated reconstruction and
direct hand sensing.
Beyond annotation quality, we conduct a preliminary exploration of training
utility through Cosmos-based~\citep{nvidia2026cosmos3} multimodal prediction
and human-to-robot motion transfer using ZeroWBC-style~\citep{zerowbc2026}
whole-body control in simulation. These initial studies suggest
benefits from combining synchronized modalities and illustrate the potential
of the recordings for policy learning. 

We summarize our contributions as follows:
\textbf{(i)} \name, a vision-motion-touch dataset that achieves rich multimodal coverage and high-fidelity motion annotations with lightweight instrumentation, capturing continuous human \locoman;
\textbf{(ii)} the $\mathrm{C}^3$ pipeline, which decouples body-sensor calibration from motion inference and combines temporal completion with layered constraints to recover metric whole-body motion from sparse trackers.
\textbf{(iii)} experiments demonstrating improved annotation quality and the effectiveness of our design choices, complemented by an initial assessment of the recordings' utility for embodied learning.

\section{Related Work}
\label{sec:related}

\noindent \textbf{Multimodal human datasets.}
Calibrated capture systems provide detailed body, hand, and interaction
geometry, but require instrumented spaces
\citep{taheri2020grab,bhatnagar2022behave,zhang2022egobody,kim2024parahome,cao2026acedata0}.
Egocentric video datasets broaden activity coverage
\citep{damen2022epic,grauman2022ego4d,grauman2024egoexo4d}, with
manipulation-oriented collections increasingly adding hand-motion annotations
\citep{hoque2025egodex,punamiya2026egoverse,li2026openaoe,li2026egolive}.
Wearable body tracking adds full-body trajectories but generally lacks direct
finger and tactile measurements
\citep{ma2024nymeria,fan2024emhi,detone2026nymeriaplus}.
Hand-focused datasets capture articulation, object geometry, or contact,
typically without the accompanying whole-body motion
\citep{banerjee2024hot3d,zhao2025egopressure,zhou2026touchanything,song2025opentouch}.
\name combines these signals in portable recordings of continuous
\locoman.

\begin{figure}[!t]
  \centering
  \includegraphics[width=0.99\linewidth]{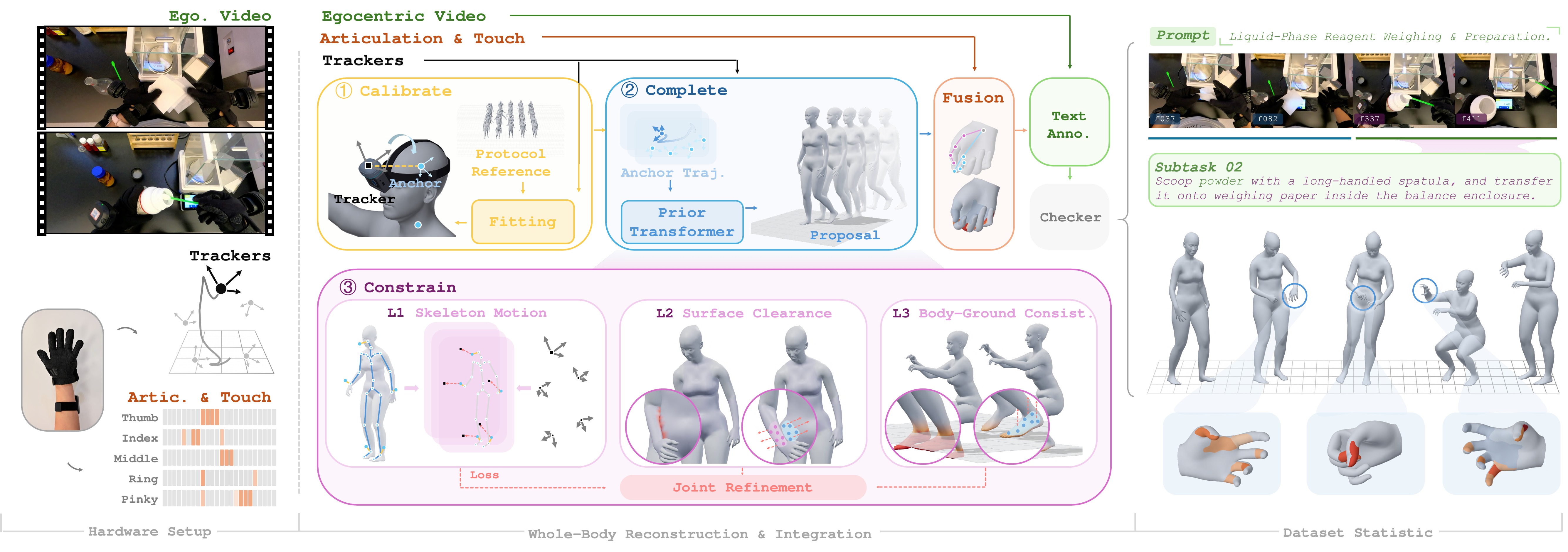}
  \caption{\textbf{Overview of \name.}
The rig records egocentric video, poses from five trackers, finger articulation, and touch (\cref{sec:system}).
$\mathbf{C}^3$ (\emph{Calibrate-Complete-Constrain}) estimates a reusable body-sensor profile through protocol-guided calibration. A temporal prior generates a whole-body motion proposal from anchors; joint refinement then promotes skeletal measurement agreement, participant-specific surface clearance, and whole-mesh ground non-penetration (\cref{sec:pipeline}).
Hand and touch are integrated with the reconstructed body, followed by expert-guided step annotation and review (\cref{sec:annotation}).
The resulting episodes align multimodal data on a shared timeline (\cref{sec:stats}).}
  \label{fig:pipeline}
  \vspace{-10pt}
\end{figure}

\noindent \textbf{Motion reconstruction from partial observations.}
Video-based reconstruction remains sensitive to occlusion, depth ambiguity,
and camera motion \citep{shen2024gvhmr,zhang2025hawor,xu2026handflow}.
Sparse-tracking methods complete motion from observations using
learned priors
\citep{jiang2022avatarposer,du2023agrol,ponton2023sparseposer,jiang2024egoposer,dai2024hmdposer}.
Kinematic fitting and physical regularization improve plausibility
\citep{jiang2022avatarposer,zheng2023avatarjlm,jiang2024manikin}.
Yet matching sparse anchors leaves intermediate joints ambiguous, and mount
errors can be absorbed into articulation.
Our $\mathrm{C}^3$ pipeline combines participant and mount calibration,
motion completion, and constrained refinement to address sensor mismatch, self-intersection, and floor penetration.

\noindent \textbf{Human data for embodied learning.}
Human demonstrations support cross-embodiment manipulation learning
\citep{kareer2024egomimic,qiu2025ph2d,punamiya2026egoverse}, mobile and humanoid
policy generalization \citep{zhu2025emma,zhong2025humanoidexo,shi2026egohumanoid},
and whole-body control through motion retargeting and contact-aware references
\citep{omniretarget2025,zerowbc2026,omnicontact2026}.
\name preserves visual context, body coordination, and motion-contact
relationships throughout continuous tasks, providing a resource for these
learning settings.

\section{The \name Dataset}
\label{sec:dataset}

Egocentric video captures task semantics at scale, yet leaves key signals for \locoman under-constrained: hand articulation is obscured;
whole-body motion, especially lower-body support, extends beyond the camera's field of view; and appearance alone cannot reveal tactile contact.
\name addresses this asymmetry by pairing a head-mounted camera with sparse wearable sensing, grounding body and hand motion in sensor measurements and corresponding touch records.
The dataset contains 2{,}715 recordings (100\hours{}) across ten task templates and five environments, spanning laboratories, factories, and homes.
Following the workflow in \cref{fig:pipeline}, we first introduce the capture hardware (\cref{sec:system}), then describe sensor calibration, body completion, and constrained refinement of fine-grained body reconstruction (\cref{sec:pipeline}), and enrich the episodes with expert-guided task boundaries and annotations (\cref{sec:annotation}).
\cref{sec:stats} offers a glimpse of the dataset.

\subsection{Capture Hardware}
\label{sec:system}

\textbf{Requirements.}
Fine-grained \locoman capture requires accurate measurements of body motion, hand articulation, and contact in a lightweight, portable system.
Studio motion capture \citep{taheri2020grab,kim2024parahome,cao2026acedata0} provides dense measurements but confines collection to a calibrated multi-camera volume, whereas readily deployable video-only capture \citep{shen2024gvhmr,zhang2025hawor} must infer occluded motion and visually ambiguous contact.
We therefore complement sparse tracking at key body joints with direct finger sensing and optional touch for whole-body reconstruction and fine-grained interaction annotation.

\textbf{Hardware.}
Five VIVE Ultimate Trackers locate the head, hands, and ankles (\cref{app:calibration}), using onboard cameras for 6-DoF inside-out tracking without external base stations.
Head and hand trackers establish metric camera and end-effector trajectories, while ankle trackers capture foot placement and constrain stance and support beyond the head camera’s view.
Custom 3D-printed mounts rigidly couple the head tracker to a GoPro HERO13 with Media Mod and attach the others to the glove dorsum and lateral ankles, with straps for added fixation.
Two interchangeable glove configurations share this layout: MANUS Metagloves measure finger articulation, while Wuji gloves additionally record tactile signals.

\textbf{Deployment.}
The compact, reusable rig simplifies sensor setup and donning, facilitating collection across environments.
Subject and mount calibration is reused across task recordings as long as the rig remains mounted.
\cref{sec:pipeline} details calibration and whole-body reconstruction from sparse measurements; \cref{app:cost-discussion} covers cost and deployment trade-offs.

\subsection{Whole-Body Reconstruction from Sparse Measurement}
\label{sec:pipeline}
\label{sec:smplx}

\textbf{$\mathbf{C}^3$: Calibrate-Complete-Constrain.}
Whole-body reconstruction calls for inferring the body joint motion from the rig's sparse tracker poses.
Yet, coupled ambiguities arise: tracker poses depend jointly on body articulation and rigid attachment to anatomical anchors (\textit{i.e.}, head, wrists, and ankle joint frames), allowing mounting errors to be absorbed into estimated motion.
Even calibrated anchors leave intermediate joints undetermined.
We address these with $\mathrm{C}^3$ (Calibrate-Complete-Constrain; \cref{fig:pipeline}), separating fixed body-sensor geometry from time-varying motion.
\emph{Protocol-guided calibration} aligns a prescribed movement sequence (\textit{i.e.}, the calibration protocol) with a reference motion library by phase.
Robust fitting estimates body shape, scale, and sensor-to-anchor transforms from these correspondences once per mounting configuration.
\emph{Prior-guided motion completion} uses a temporal motion prior, a model for completing unobserved articulation, to generate a motion proposal: an initial whole-body trajectory from anchors.
\emph{Layered constraints} then jointly refine skeletal measurement agreement, surface clearance, and ground non-penetration.
We fuse glove-measured finger articulation and touch as direct interaction evidence.

\subsubsection{Protocol-guided calibration}
\label{sec:lightweight-calibration}

\textbf{Calibration problem definition.}
Calibration estimates a participant-and-mount profile
$\Phi=(\beta,s,\{A_i,a_i\}_{i=1}^{5})$.
Here $\beta\in\mathbb R^{10}$ specifies \smplx body shape~\citep{pavlakos2019smplx} and $s>0$
converts model dimensions to the participant's metric dimensions.
For tracker $i$, $a_i\in\mathbb R^3$ is the displacement from
the anchor to the device origin. $A_i\in\mathrm{SO}(3)$ expresses the device's axes in that same
frame, where $\mathrm{SO}(3)$ denotes proper 3D rotation matrices.
The five attachments correspond to the head, wrists, and ankles.
These parameters remain shared across calibration actions and subsequent
tasks with unchanged mounts; joint articulation varies.
We estimate $\Phi$ by anchoring body dimensions, matching action phases,
and fitting one attachment per device across the matched poses.

\textbf{Body shape and scale initialization.}
Visual evidence provides an initial estimate of body proportions; direct
height measurement anchors its metric size. We obtain
$\beta^{\mathrm{vis}}\in\mathbb R^{10}$ from a short participant video using
GVHMR~\citep{shen2024gvhmr} and measure the participant's height $h^\star$.
Let $H(\beta)$ be the unscaled model height in neutral pose, then $\beta_0=\beta^{\mathrm{vis}}$ initializes the shape fit.
Taking the visual shape as our starting point, we rescale it to the
measured height $s_0=\frac{h^\star}{H(\beta^{\mathrm{vis}})}$.
The resulting $(\beta_0,s_0)$ initializes the geometry fit below.
The measurements are collected for the reusable profile.

\textbf{Protocols.}
To disentangle body proportions, sensor offsets, and world-frame alignment, we use a calibration protocol comprising standing, A- and T-poses, forward arm extension, head and wrist rotations, elbow flexion, shallow squats, leg raises, and ankle motion. A reference motion library provides consistent anatomical targets without requiring an exact match to the participant's motion. To accommodate differences in execution speed, we automatically match ordered holds and motion peaks in recorded tracker signals to reference actions, pairing observed device poses with corresponding reference joint poses. The library is cached and reused, limiting participant-specific work to protocol recording and profile fitting. The full protocol and sample-selection rules are detailed in \cref{app:calibration-method}.

\textbf{Forward model for calibration.}
A tracker's pose is jointly determined by the body shape, the pose locating the anchor, and the attachment placing
the device relative to it.
Let $n=1,\ldots,N_{\mathrm{cal}}$ index matched calibration samples, with $r(n)$ identifying the recording containing sample $n$.
Each sample corresponds to reference action $k_n$ at normalized phase $\alpha_n\in[0,1]$ ($0$: start; $1$: end).
Phase-based correspondence accommodates differences in execution speed, and each action may contribute multiple samples.
For body shape $\beta$, the matched reference pose specifies anchor $i$'s pelvis-relative position $J_{n,i}(\beta)=J_i^{k_n}(\alpha_n;\beta)$ and orientation $G_{n,i}=G_i^{k_n}(\alpha_n)$ in the reference body frame.
Let $s$ denote body scale, and let $a_i$ and $A_i$ denote the fixed attachment translation and rotation expressed in anchor $i$'s frame.
Recording-specific yaw rotation $Q_r$ and pelvis translation $u_r$ then give the predicted tracker position and orientation in the tracking world frame:
\begin{equation}
\widehat p_{n,i}=u_{r(n)}
+Q_{r(n)}\!\left[sJ_{n,i}(\beta)+G_{n,i}a_i\right],\
\widehat R_{n,i}=Q_{r(n)}G_{n,i}A_i .
\label{eq:calibration-forward}
\end{equation}
Here $\widehat p_{n,i}$ and
$\widehat R_{n,i}$ are the predicted position and orientation of tracker $i$ at matched sample $n$.
The term $sJ_{n,i}(\beta)$ locates the joint at metric scale and $G_{n,i}a_i$
adds the mounting offset, $\widehat p_{n,i}$ enters the position loss to fit profile term $\Theta$;
the relation for $\widehat R_{n,i}$ is then used separately to recover $A_i$.

\textbf{Joint calibration objective.}
We define a calibration loss to jointly estimate body geometry,
attachment displacements, and recording alignments, collected in
$\Theta=(\beta,s,\{a_i\}_{i=1}^{5},\{Q_r,u_r\}_r)$.
Anthropometric measurements and geometric priors constrain this fit to discourage compensation between limb dimensions and mounting offsets.
Initializing $(\beta,s)$ at $(\beta_0,s_0)$, we solve:
\begin{equation}
\Theta^{\mathrm{fit}}=\arg\min_{\Theta:\,s>0}\mathcal L_{\mathrm{cal}}(\Theta),\
\mathcal L_{\mathrm{cal}}(\Theta)=
\lambda_{\mathrm{obs}}\mathcal L_{\mathrm{robust\mbox{-}pos}}
+\lambda_{\mathrm{anth}}\mathcal L_{\mathrm{anth}}
+\mathcal R_{\mathrm{geom}} .
\label{eq:calibration-objective}
\end{equation}
Here $\Theta^{\mathrm{fit}}$ denotes the fitted parameters and the non-negative $\lambda$ coefficients weight the loss terms.
$\mathcal L_{\mathrm{robust\mbox{-}pos}}$ compares predicted tracker
positions $\widehat p_{n,i}$ from \cref{eq:calibration-forward} with
tracker-measured positions $p_{n,i}$;
$\mathcal L_{\mathrm{anth}}$ constrains height and arm span to their
measurements; and $\mathcal R_{\mathrm{geom}}$ anchors shape to the visual
estimate and regularizes attachments, scale, and yaw.
Reference rotations remain fixed during optimization.
Loss definitions, squat-translation handling, and solver settings are in
\cref{app:calibration-method}.

\textbf{Attachment rotation recovery.}
We next estimate the remaining rotation $A_i$ by matching
predicted tracker orientations $\widehat R_{n,i}$ to measured ones
$R_{n,i}$. With $Q_{r(n)}$ and $G_{n,i}$ fixed, the rotation-calibration
routine $f_{\mathrm{rot}}$ fits one shared $A_i$ through the orientation
model in \cref{eq:calibration-forward} across matched samples.
We summarize this fitting step as
$A_i=f_{\mathrm{rot}}(\{\widehat R_{n,i},R_{n,i}\}_n)$, completing $\Phi$ (\cref{app:calibration-method}).

\subsubsection{Prior-guided motion completion}
\label{sec:body-hand-reconstruction}

\textbf{Reconstruction requirements.}
With the calibrated profile $\Phi$ fixed, we reconstruct $\mathcal X=\{x_t\}_{t=1}^{T}$, where $x_t=(\omega_t,\theta_t,u_t)$ comprises the global pelvis orientation, the local rotations of the 21 \smplx joints relative to their respective parents, and the pelvis position in the world frame.
Joint positions follow from forward kinematics with the calibrated body shape $\beta$ and scale $s$.
A temporal motion prior completes unobserved articulation from the five anchors, yielding an initial motion proposal.
A proposal that is plausible under the prior may still deviate from calibrated anchor poses or exhibit body self-interpenetration and ground penetration.
We therefore refine it under three constraint layers: skeletal motion, participant-specific surface clearance, and body-ground consistency.

\textbf{Motion completion.}
The calibrated attachments map measured tracker poses $(p_{t,i},R_{t,i})$ to anatomical anchor frames: $\widetilde R_{t,i}=R_{t,i}A_i^\top$ and $\widetilde p_{t,i}=p_{t,i}-\widetilde R_{t,i}a_i$. These frames locate the head, wrists, and ankles but leave intermediate articulation underdetermined. To complete the unobserved motion, we extend AvatarPoser~\citep{jiang2022avatarposer} from head-and-hand inputs to all five anchors and train the transformer on AMASS~\citep{mahmood2019amass}. The resulting motion prior $f_{\mathrm{prior}}$ maps an $L$-frame observation history to initial root orientation and local body-joint rotations: $(\omega_t^0,\theta_t^0)=f_{\mathrm{prior}}(\mathbf z_{t-L+1:t})$, where $\mathbf z_t$ encodes calibrated positions, rotations, and their temporal increments for all five anchors. We recover the pelvis position by aligning the predicted head position with its calibrated anchor: $u_t^0=\widetilde p_{t,\mathrm{head}}-s\bar J_{\mathrm{head}}(\omega_t^0,\theta_t^0;\beta)$, where $\bar J_{\mathrm{head}}$ is the unscaled pelvis-to-head displacement under the predicted root orientation and joint rotations. Together, these estimates form the initial trajectory $\mathcal X^0=\{(\omega_t^0,\theta_t^0,u_t^0)\}_{t=1}^{T}$. Model training is detailed in \cref{app:proposal-training}.

\begin{figure}[!t]
  \centering
  \includegraphics[width=\linewidth]{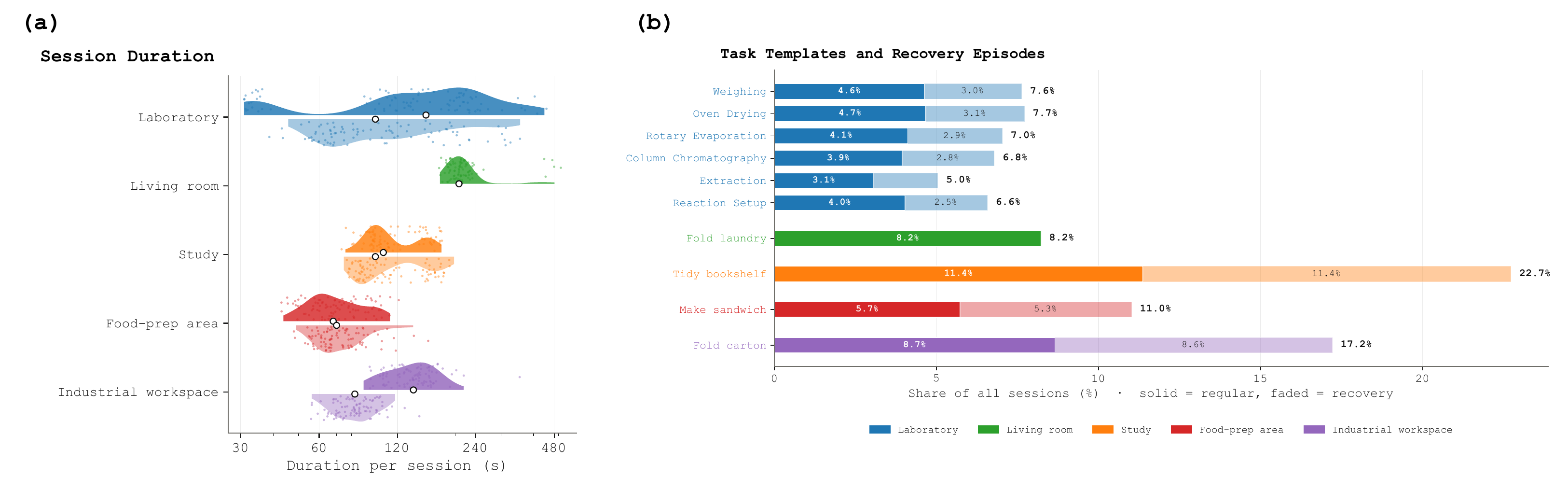}
  \caption{\textbf{Data statistics}. \textbf{(a)} Session durations;
  \textbf{(b)} regular and recovery recordings across 10 tasks.}
  \label{fig:dataset-overview}
  \label{fig:task-statistics}
  \vspace{-10pt}
\end{figure}

\subsubsection{Layered constraints}
\label{sec:layered-constraints}

\textbf{Constraint layer 1: skeletal motion.}
The learned completion provides an initial articulation; refinement fits it
to the current recording's measurements.
With shape, scale, and attachments fixed, forward kinematics maps each candidate state $x_t$ to predicted tracker poses $(\widehat p_{t,i},\widehat R_{t,i})$ through \cref{eq:calibration-forward}. 
Position and orientation losses $\mathcal L_p$ and $\mathcal L_R$ compare these predictions with valid raw tracker observations. 
To regularize the fit, $\mathcal L_{\mathrm{prior}}$ discourages unsupported deviations from $\mathcal X^0$, $\mathcal L_{\mathrm{time}}$ regularizes temporal changes in articulation, and $\mathcal L_{\mathrm{joint}}$ penalizes extreme poses through soft joint limits. Together, these terms define $\mathcal L_{\mathrm{skel}}=\sum_k\lambda_k\mathcal L_k$, with details in \cref{app:skeletal-losses}.

\textbf{Constraint layer 2: surface clearance.}
Skeletal agreement does not account for body thickness: an arm may intersect the torso even when its wrist matches the target. We therefore constrain the posed \smplx surface using the participant's calibrated shape and scale. For nearby points sampled from distinct body parts, $\mathcal L_{\mathrm{surf}}$ penalizes insufficient oriented clearance along the target surface normal, distinguishing points outside the surface from those that have crossed it. Aggregating the largest violations per part pair concentrates corrections in problematic regions, discouraging self-intersection while respecting participant-specific geometry (\cref{app:surface-loss}).

\textbf{Constraint layer 3: body-ground consistency.}
An ankle anchor above the floor does not ensure that the sole, heel, or knee
also clears it. We apply $\mathcal L_{\mathrm{floor}}$ to all posed body
vertices, penalizing their depth below the estimated ground plane. This
constrains the body's placement relative to the environment, including
regions that approach the floor during crouching or kneeling. The term
penalizes penetration without imposing foot-contact or foot-locking constraints.

\textbf{Joint refinement.}
The three layers act on the same trajectory through a shared objective:
\begin{equation}
 \mathcal X^\star=\arg\min_{\mathcal X}
 \mathcal L_{\mathrm{body}}(\mathcal X;\Phi),\
 \mathcal L_{\mathrm{body}}(\mathcal X;\Phi)=
 \mathcal L_{\mathrm{skel}}
 +\lambda_{\mathrm{surf}}\mathcal L_{\mathrm{surf}}
 +\lambda_{\mathrm{floor}}\mathcal L_{\mathrm{floor}}.
 \label{eq:body-objective}
\end{equation}
The non-negative $\lambda$ coefficients weight the surface and ground terms;
$\mathcal L_{\mathrm{skel}}$ includes its component weights.
Starting from $\mathcal X^0$, the solver evaluates all three layers at every
iteration and differentiates through \smplx forward kinematics and skinning
to update $(\omega_t,\theta_t,u_t)$, keeping $\Phi$ fixed. Optimization runs
in temporal chunks, whose fitted states form $\mathcal X^\star$ (\cref{app:body-solver}).

\par

\subsection{Multimodal integration and annotation}
\label{sec:annotation}
\label{sec:validation}

\textbf{Hand and touch fusion.}
We directly fuse glove-measured finger geometry to the \smplx hands in wrist-local
coordinates.
Tactile observations are aligned in time and associated with the
corresponding hand regions. Hand fitting and
tactile processing are detailed in \cref{app:hand-fusion}.

\textbf{Expert-guided procedural context.}
We condition off-the-shelf VLM tools~\citep{peng2026cortex} on egocentric video, task instructions, and expert-provided skill documents outlining workflow stages, typical progression, and relevant equipment and operations (\cref{app:text-skills}). The tools produce subtask segments with start/end times, skill labels, object/tool roles, observed outcomes, and concise descriptions; uncertain outcomes remain unspecified. Video timestamps align the annotations with motion and tactile streams. Finally, operators manually review video, reconstructed motion, available touch, and text for temporal alignment and semantic consistency.



\section{Data Statistics}
\label{sec:stats}
\label{sec:tasks}

\begin{figure}[!t]
  \centering
  \includegraphics[width=0.99\linewidth]{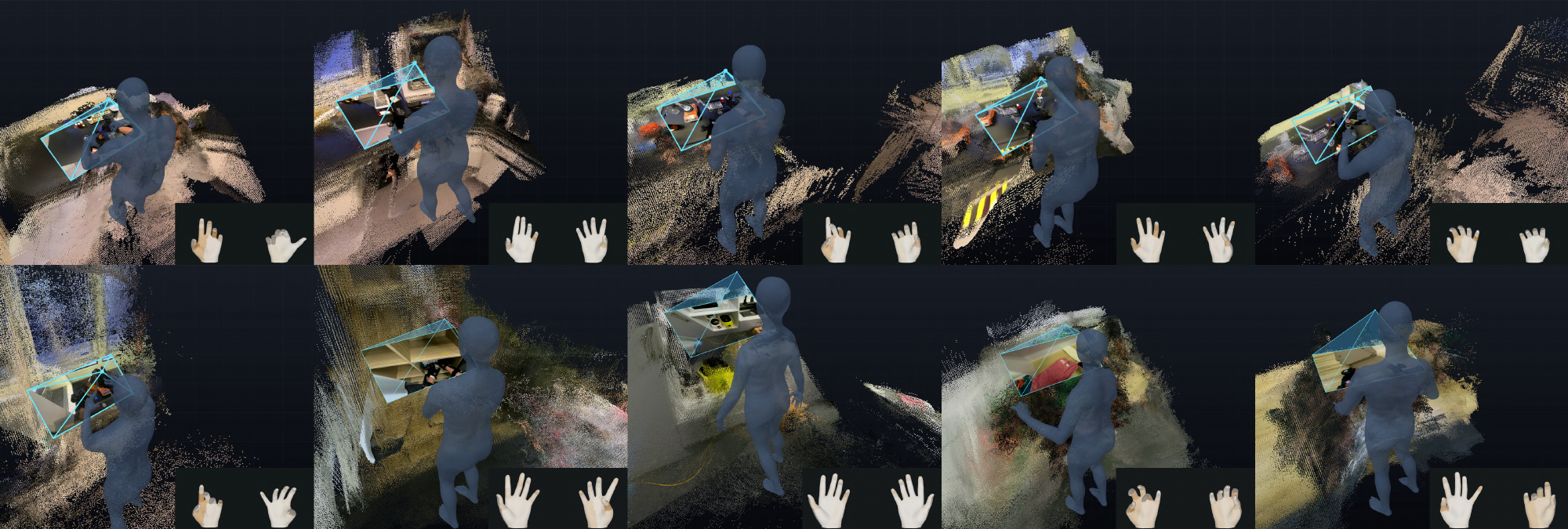}
  \caption{\textbf{Synchronized body motion and touch across ten tasks.}
Each panel shows a matched instant, with tactile readings at lower right; gray-to-orange shading indicates baseline-relative increases.}
  \label{fig:scene-body-touch}
  \vspace{-10pt}
\end{figure}

\begin{figure}[!t]
  \centering
  \includegraphics[width=\textwidth]{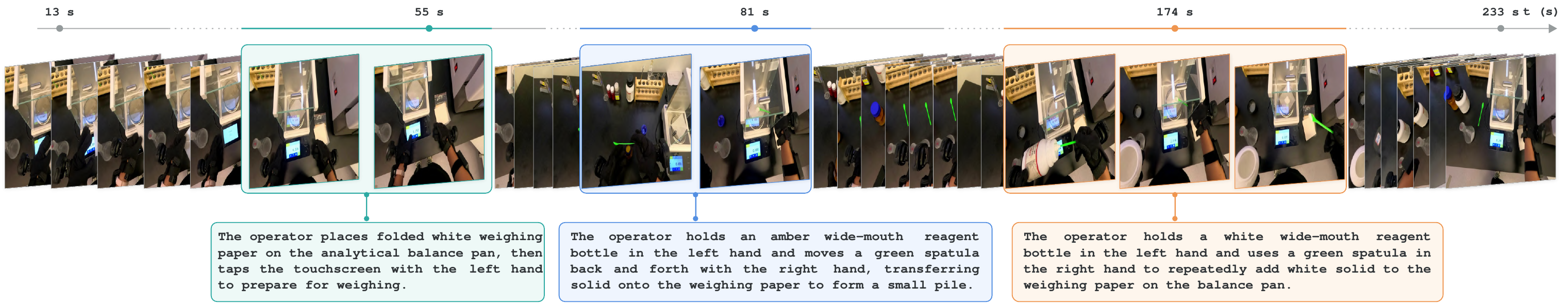}
  \caption{\textbf{Step annotations.}
  Three color-coded subtasks with timestamps and descriptions.}
  \label{fig:text-annotation-visual}
  \vspace{-10pt}
\end{figure}

\name comprises 2{,}715 recordings (100 hours) spanning ten task templates across five environments, with an emphasis on professional activities.
\cref{fig:dataset-overview} summarizes session duration and the balance
between regular and recovery episodes, including responses to interruptions
or failed attempts. \cref{fig:scene-body-touch,fig:text-annotation-visual}
illustrate synchronized body--touch observations and procedural text.
Further statistics on sensing coverage, motion, contact, text, and
reconstructed environments are provided in \cref{app:stats}.

\newcommand{\evalpending}{\textcolor{orange!85!black}{\textbf{--}}}
\newcommand{\evalna}{\textcolor{gray}{n/a}}
\section{Experiments}
\label{sec:experiments}
We organize evaluation around annotation quality, preliminary training
utility, and key design choices.

\textbf{Evaluation setup.}
Our purpose-built \textbf{\evalsubset} covers whole-body motion, hand
articulation, and regional contact. Evaluation addresses three aspects:
\emph{kinematic accuracy}, through end-point and full-body point errors;
\emph{temporal quality}, through tracking continuity and trajectory smoothness;
and \emph{contact accuracy}, through regional IoU and F1. References are
modality-specific, with full-body errors measured against tracker-fitted labels.
The remaining \textbf{\naturaldata} lack matched accuracy references and are assessed for sensor agreement, temporal quality, and geometric consistency.
Subsets, metrics, and full-body references are detailed in \cref{app:evaluation-data,app:evaluation-metrics,app:body-reference}.


\subsection{Annotation Reliability}
\label{sec:exp-reliability}
\textbf{Whole-body reconstruction.}
\label{sec:exp-body}\label{sec:bench-body}
\cref{tab:recon} compares three baseline settings: public pretrained
checkpoints with calibrated sensor-to-anchor conversion, direct use of raw
tracker poses, and fine-tuning with five tracker anchors. \name outperforms
all three settings across every reported metric, achieving 18.0\,mm body-point
error and 20.3\,mm at non-tracked points. Its lowest errors extend beyond the
observed anchors to other body points, including elbows and knees, indicating
better recovery of full-body articulation.
Visual comparisons are provided in \cref{fig:recon-matched}.

\begin{table}[!t]
\centering\footnotesize
\caption{\textbf{Body reconstruction on the Evaluation Subset (mm; velocity: mm/s).} Non-tracked: 17 unobserved joints.
$^\dagger$Reconstructed pelvis; $^\ddagger$raw tracker poses;
$^*$locally trained variants.}
\label{tab:recon}
\setlength{\tabcolsep}{2pt}
\scriptsize
\begin{tabular*}{\linewidth}{@{\extracolsep{\fill}}lcccccc@{}}
\toprule
Method & Anchors & Body points\dn & Non-tracked\dn &
PA-MPJPE\dn & Root\dn & Velocity\dn\\
\midrule
AvatarPoser~\citep{jiang2022avatarposer} & 3 & 128.5 & 122.5 & 86.3 & 99.5 & 106.1\\
AGRoL~\citep{du2023agrol} & 3 & 119.6 & 114.0 & 81.0 & 89.6 & 113.0\\
EgoPoser~\citep{jiang2024egoposer} & 3 & 155.5 & 143.2 & 89.3 & 121.2 & 113.2\\
AvatarJLM~\citep{zheng2023avatarjlm} & 3 & 117.9 & 113.3 & 78.6 & 95.0 & 105.9\\
SAGE~\citep{feng2024sage} & 3 & 114.7 & 109.5 & 81.4 & 83.4 & 110.3\\
SparsePoser~\citep{ponton2023sparseposer}$^\dagger$ & 6$^\dagger$ & 219.2 & 189.9 & 134.4 & 84.3 & 119.1\\
\midrule
AvatarPoser$^\ddagger$ & 3 & 235.3 & 239.9 & 93.0 & 251.1 & 109.6\\
AGRoL$^\ddagger$ & 3 & 227.5 & 230.0 & 88.7 & 232.0 & 148.8\\
AvatarPoser$^*$ & 5 & 64.1 & 62.1 & 21.0 & 61.7 & 96.8\\
HMD-Poser~\citep{dai2024hmdposer}$^*$ & 5 & 136.7 & 131.5 & 71.3 & 107.8 & 152.1\\
\midrule
\name & 5 & \textbf{18.0} & \textbf{20.3} & \textbf{13.3} & \textbf{25.9} & \textbf{83.0}\\
\bottomrule
\end{tabular*}
\end{table}

\begin{table}[!t]
\centering
\begin{minipage}[t]{0.565\linewidth}
\centering\footnotesize
\caption{\textbf{Hand capture quality on the \evalsubset.}
Tip/Hold RMS: mm; Missing/Success: \%; Gap: s.}
\label{tab:hand-quality}
\vspace{-10pt}
\setlength{\tabcolsep}{1.2pt}
\scriptsize
\begin{tabular}{@{}lccccc@{}}
\toprule
Method & Tip\dn & Hold RMS\dn & Missing\dn & Gap p95\dn & Success\up\\
\midrule
HaWoR~\citep{zhang2025hawor} & 27.1 & 4.5 & 7.9 & 2.13 & 47\\
HandFlow~\citep{xu2026handflow} & 30.7 & 11.3 & 7.8 & 2.09 & 67\\
\textbf{Ours} & \textbf{9.5} & \textbf{0.4} & \textbf{0.0} & \textbf{0.00} & \textbf{100}\\
\bottomrule
\end{tabular}
\end{minipage}\hfill
\begin{minipage}[t]{0.405\linewidth}
\centering\footnotesize
\caption{\textbf{Contact quality on the \evalsubset using frame-region labels.}}
\label{tab:tactile}
\vspace{-10pt}
\setlength{\tabcolsep}{4pt}
\scriptsize
\begin{tabular}{@{}lcc@{}}
\toprule
Input & IoU\up & F1\up\\
\midrule
RGB (HACO~\citep{jung2025haco}) & 0.621 & 0.766\\
Hand pose only & 0.324 & 0.490\\
\textbf{Ours} & \textbf{0.915} & \textbf{0.956}\\
\bottomrule
\end{tabular}
\end{minipage}
\vspace{-10pt}
\end{table}

\begin{table}[!t]
\centering\footnotesize
\caption{\textbf{Expert review of model-generated text annotations (\%).}}
\vspace{-10pt}
\label{tab:language}
\setlength{\tabcolsep}{3pt}
\scriptsize
\begin{tabular*}{\linewidth}{@{\extracolsep{\fill}}lcccccc@{}}
\toprule
Annotation & Fact error\dn & Unsupported\dn & Missed steps\dn &
Extra steps\dn & Boundary error\dn & Unjudgeable\dn\\
\midrule
Model-generated & 11.09 & 2.46 & 8.16 & 2.05 & 3.08 & 0.82\\
\bottomrule
\end{tabular*}
\vspace{-10pt}
\end{table}

\textbf{Hand motion and touch capture.}
\label{sec:exp-tactile}\label{sec:bench-tactile}
\cref{tab:hand-quality,tab:tactile} compare our glove settings
with visual and kinematic inference. In matched-configuration hand tests,
our glove setting yields 0.4\,mm holding RMS, versus 4.5\,mm for
HaWoR~\citep{zhang2025hawor} and 11.3\,mm for HandFlow~\citep{xu2026handflow},
and achieves 100\% kinematic target recovery.
For regional contact, our glove setting achieves 0.915 IoU,
compared with 0.621 for HACO~\citep{jung2025haco} and 0.324 for hand-pose-only
inference. Together, these tests support direct sensing of finger articulation
and contact.

\textbf{Step text quality.}
\label{sec:exp-lang}\label{sec:bench-lang}
Expert review of model-generated annotations (\cref{tab:language}) finds
factual errors in 11.09\% of reviewed segments and unsupported claims in
2.46\%. The audit also reports missed- and extra-step rates of 8.16\% and
2.05\%, respectively, with boundary corrections in 3.08\% of segments.
These residual errors motivate manual verification
and correction in our workflow before release.

\begin{table}[!t]
\centering\footnotesize
\begin{minipage}[c]{0.47\linewidth}
\centering
\begin{minipage}[c]{0.556\linewidth}
\centering\scriptsize
\setlength{\tabcolsep}{1pt}
\begin{tabular*}{\linewidth}{@{\extracolsep{\fill}}lcc@{}}
\toprule
Metric & Separate & Joint\\
\midrule
Video PSNR (dB)\up & 15.94 & \textbf{16.62}\\
Action MSE\dn & 0.059 & \textbf{0.044}\\
Contact F1\up & 0.610 & \textbf{0.694}\\
\bottomrule
\end{tabular*}
\end{minipage}\hfill
\begin{minipage}[c]{0.422\linewidth}
\centering\scriptsize
\setlength{\tabcolsep}{1pt}
\begin{tabular*}{\linewidth}{@{\extracolsep{\fill}}lc@{}}
\toprule
Condition & Action MSE\dn\\
\midrule
None & 0.0478\\
Video & 0.0277\\
Video+touch & \textbf{0.0264}\\
\bottomrule
\end{tabular*}
\end{minipage}
\end{minipage}\hfill
\begin{minipage}[c]{0.52\linewidth}
\centering
\includegraphics[width=\linewidth]{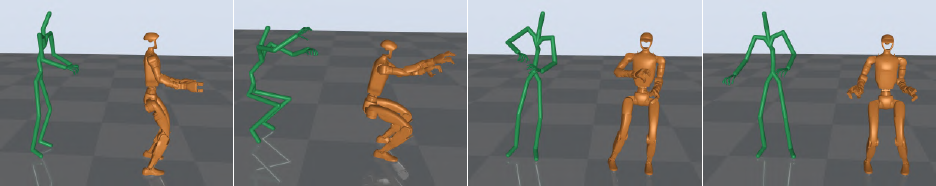}
\end{minipage}
\par\noindent
\begin{minipage}[t]{0.47\linewidth}
\vspace{0pt}
\captionsetup{position=bottom,skip=6pt}
\caption{\textbf{Preliminary study.} \textbf{(a)} Separate/joint generation;
\textbf{(b)} conditional action generation.}
\label{tab:downstream}
\end{minipage}\hfill
\begin{minipage}[t]{0.52\linewidth}
\vspace{0pt}
\captionsetup{position=bottom,skip=6pt}
\captionof{figure}{\textbf{Human motion (green) and G1 whole-body control policy
in simulation (gold).}}
\label{fig:humanoid-control}
\end{minipage}
\vspace{-10pt}
\end{table}

\begin{table}[!t]
\centering\scriptsize
\caption{\textbf{Ablations on the \evalsubset and \naturaldata.}
Pose errors: mm; velocity/slide: mm/s; jerk: m/s$^3$; floor: \%. Blue: primary metric (Avg.); gray: the baseline.}
\label{tab:ablation}
\vspace{-10pt}
\setlength{\tabcolsep}{1.5pt}
\scriptsize
\settowidth{\summarycolwidth}{Avg.\dn}
\addtolength{\summarycolwidth}{2\tabcolsep}
\begin{tikzpicture}
\node[inner sep=0pt, outer sep=0pt] (ablationtable) {%
\begin{tabular*}{\linewidth}{@{\extracolsep{\fill}}l*{13}{c}@{}}
\toprule
\multirow{2}{*}[-2.5pt]{Configuration} & \multicolumn{3}{c}{Layers} & \multicolumn{4}{c}{\evalshort} & \multicolumn{5}{c}{\naturalshort} & \multirow{2}{*}[-2.5pt]{Avg.\dn}\\
\cmidrule(lr){2-4}\cmidrule(lr){5-8}\cmidrule(lr){9-13}
& L1 & L2 & L3 & Body pts\dn & Non-trk.\dn & PA\dn & Velocity\dn & Tracker\dn & Crossing\dn & Floor\dn & Jerk & Slide\dn & \\
\midrule
\textit{Full (1+2+3)} & \cmark & \cmark & \cmark & \textit{18.0} & \textit{20.3} & \textit{13.3} & \textit{83.0} & \textit{10.5} & \textit{38.1} & \textit{6.0} & \textit{37.6} & \textit{29.1} & \textit{\textbf{28.4}}\\
\midrule
Full, mean body & \cmark & \cmark & \cmark & 59.5 & 66.3 & 27.3 & 96.9 & 12.3 & 71.4 & 50.9 & 49.0 & 37.5 & 52.3\\
Full, no mount calibration & \cmark & \cmark & \cmark & 95.0 & 96.6 & 33.7 & 91.6 & 41.4 & 41.3 & 3.5 & 46.0 & 52.3 & 55.7\\
Full, uncalib. at test & \cmark & \cmark & \cmark & 98.1 & 99.6 & 38.8 & 97.3 & 27.5 & 23.1 & 0.6 & 51.0 & 54.9 & 54.5\\
Layer 1 & \cmark &  &  & 16.1 & 18.2 & 11.9 & 81.4 & 9.9 & 72.8 & 57.3 & 33.5 & 27.1 & 36.5\\
Layer 1+2 & \cmark & \cmark &  & 18.1 & 20.4 & 13.2 & 82.3 & 10.1 & 38.7 & 61.4 & 34.4 & 28.7 & 34.1\\
Layer 1+3 & \cmark &  & \cmark & 15.9 & 18.0 & 11.9 & 82.0 & 10.3 & 73.0 & 5.9 & 36.6 & 27.3 & 31.2\\
\textcolor{black!45}{Proposal only} &  &  &  & \textcolor{black!45}{79.0} & \textcolor{black!45}{75.1} & \textcolor{black!45}{29.2} & \textcolor{black!45}{125.6} & \textcolor{black!45}{109.6} & \textcolor{black!45}{82.6} & \textcolor{black!45}{29.5} & \textcolor{black!45}{19.8} & \textcolor{black!45}{117.6} & \textcolor{black!45}{74.2}\\
\bottomrule
\end{tabular*}};
\begin{scope}[on background layer]
\fill[blue!6] ([xshift=-\summarycolwidth]ablationtable.north east)
  rectangle (ablationtable.south east);
\end{scope}
\end{tikzpicture}
\vspace{-20pt}
\end{table}

\begin{figure}[!t]
\centering
\includegraphics[width=\linewidth]{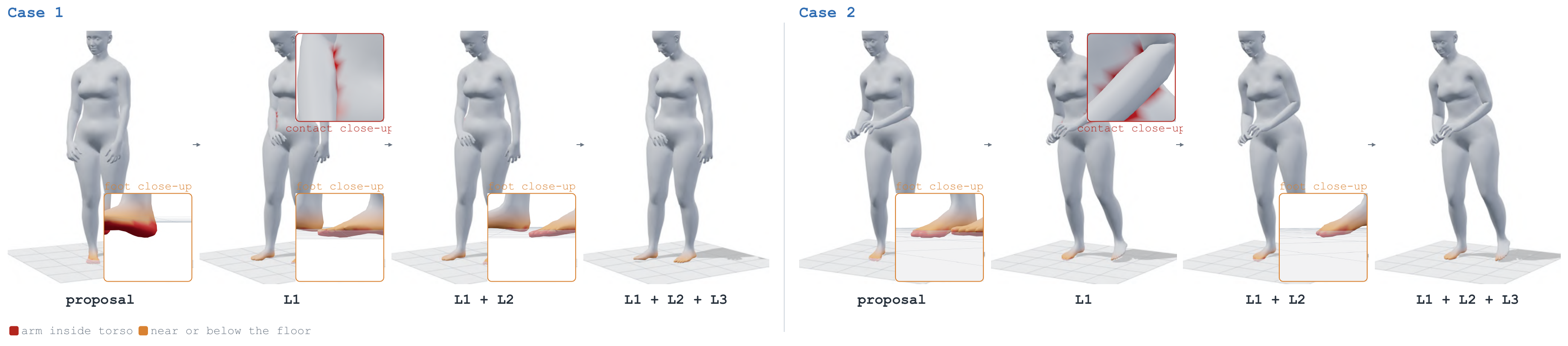}
\vspace{-20pt}
\caption{\textbf{Geometric refinement reduces self-intersection and floor
penetration.}}
\label{fig:recon}
\vspace{-10pt}
\end{figure}

\subsection{Preliminary Training Study}
\label{sec:exp-gen}\label{sec:bench-gen}
We adapt Cosmos3-Edge~\citep{nvidia2026cosmos3} to generate video,
SMPL-X-derived action, and touch
(\cref{tab:downstream}). With common video/action/touch history, we compare
separate and joint generation using sampled predictions at 5K training
steps. Joint generation improves video PSNR from 15.94 to 16.62\,dB and
contact F1 from 0.610 to 0.694 over the separate models, with a lower action MSE.
Conditioning action generation on recorded video over the
prediction interval reduces action MSE from 0.0478 to 0.0277;
adding touch reduces it to 0.0264. These conditions reveal complementary
information across modalities. Retargeting two recordings to a Unitree G1
and tracking them with a ZeroWBC-style~\citep{zerowbc2026} whole-body controller
provides a preliminary demonstration of human-to-robot motion transfer in
simulation (\cref{fig:humanoid-control}).
These initial studies illustrate multimodal learning and simulated motion
tracking; protocols are detailed in \cref{app:future-prediction,app:retarget}.

\subsection{Ablation Study}
\label{sec:ablation}\label{sec:exp-construction}
\textbf{Calibration and layered reconstruction.}
\cref{tab:ablation} shows complementary gains from successively adding Layers~1--3: body-point error drops from 79.0 to 16.1\,mm, crossing pairs from 72.8 to 38.7 per frame, and floor violations from 61.4\% to 6.0\%, respectively. Full has the lowest Avg. (28.4; mean of nine reported values), despite higher body-point error and jerk than Layer~1 alone. Replacing calibrated shape and scale with the mean body raises body-point error from 18.0 to 59.5\,mm and floor violations from 6.0\% to 50.9\%. Omitting mount calibration during training raises body-point error to 95.0\,mm, while omitting it only at test time yields 98.1\,mm. These results support fixing body-sensor geometry before motion recovery. \cref{fig:recon} shows reduced self-intersection and floor penetration.

\section{Conclusion}
\label{sec:conclusion}


This paper introduces \name, an egocentric dataset aligning whole-body motion, hand articulation, contact, and descriptions for human \locoman. 
The proposed $\mathrm{C}^3$ pipeline recovers metric body motion from sparse wearable sensing through calibration, temporal completion, and layered constraints. 
Evaluations show the benefits of direct sensing, and support this portable approach to capturing and studying coordinated movement and manipulation.

\noindent\textbf{Limitations.}
Reconstruction cannot guarantee exact recovery of all joints, while tactile sensing has variable reliability. Preliminary studies leave benefits for real-world humanoid learning unverified.
\section*{Ackonwledgement}
This project is supported by Shanghai Artificial Intelligence Laboratory.
We are grateful to all data collection contributors for their dedicated efforts in large-scale data acquisition.
We also acknowledge the support and guidance from Prof. Minyi Guo and Prof. Hongzi Zhu. 


\bibliography{egoexplorer}

\newpage
\appendix
\raggedbottom
\makeatletter
\let\@addtocurcol\ego@standardaddtocurcol
\setlength{\@fptop}{0pt}
\setlength{\@fpsep}{10pt plus 2pt}
\setlength{\@fpbot}{0pt plus 1fil}
\makeatother
\setcounter{topnumber}{4}
\setcounter{totalnumber}{6}
\renewcommand{\topfraction}{0.95}
\renewcommand{\textfraction}{0.05}
\renewcommand{\floatpagefraction}{0.75}
\setlength{\floatsep}{10pt plus 2pt minus 2pt}
\setlength{\textfloatsep}{12pt plus 2pt minus 2pt}
\setlength{\intextsep}{6pt plus 2pt minus 1pt}

\section*{Appendix Contents}
\pdfbookmark[0]{Appendix Contents}{appendix-contents}
\begingroup
\small
\setlength{\parindent}{0pt}
\setlength{\parskip}{0pt}
\newcommand{\appindexsection}[1]{%
  \par\addvspace{6pt}%
  \noindent\hyperref[#1]{\textbf{\makebox[2.1em][l]{\ref*{#1}}\nameref*{#1}}}%
  \nobreak\hspace{0.6em}\dotfill\hspace{0.6em}\hyperref[#1]{\pageref*{#1}}\par}
\newcommand{\appindexsubsection}[1]{%
  \par\addvspace{2pt}%
  \noindent\hspace*{1.1em}\hyperref[#1]{\makebox[2.9em][l]{\ref*{#1}}\nameref*{#1}}%
  \nobreak\hspace{0.6em}\dotfill\hspace{0.6em}\hyperref[#1]{\pageref*{#1}}\par}
\appindexsection{app:calibration}
\appindexsection{app:schema}
\appindexsection{app:reconstruction}
\appindexsubsection{app:calibration-method}
\appindexsubsection{app:proposal-training}
\appindexsubsection{app:skeletal-losses}
\appindexsubsection{app:surface-loss}
\appindexsubsection{app:ground-loss}
\appindexsubsection{app:body-solver}
\appindexsection{app:integration}
\appindexsubsection{app:hand-fusion}
\appindexsubsection{app:skill-structure}
\appindexsubsection{app:skill-prompt}
\appindexsubsection{app:skill-review}
\appindexsection{app:stats}
\appindexsection{app:ablation}
\appindexsubsection{app:evaluation-data}
\appindexsubsection{app:body-reference}
\appindexsubsection{app:glove-evaluation}
\appindexsubsection{app:evaluation-metrics}
\appindexsubsection{app:review-protocol}
\appindexsubsection{app:future-prediction}
\appindexsubsection{app:reconstruction-ablation}
\appindexsubsection{app:additional-experiments}
\appindexsubsection{app:retarget}
\appindexsubsection{app:reconstruction-experiments}
\appindexsection{app:cost-discussion}
\endgroup
\clearpage

\section{Capture Hardware}
\label[appendix]{app:calibration}
\afterpage{\begin{figure}[!tp]
\centering
\includegraphics[width=0.99\linewidth,height=0.80\textheight,keepaspectratio]{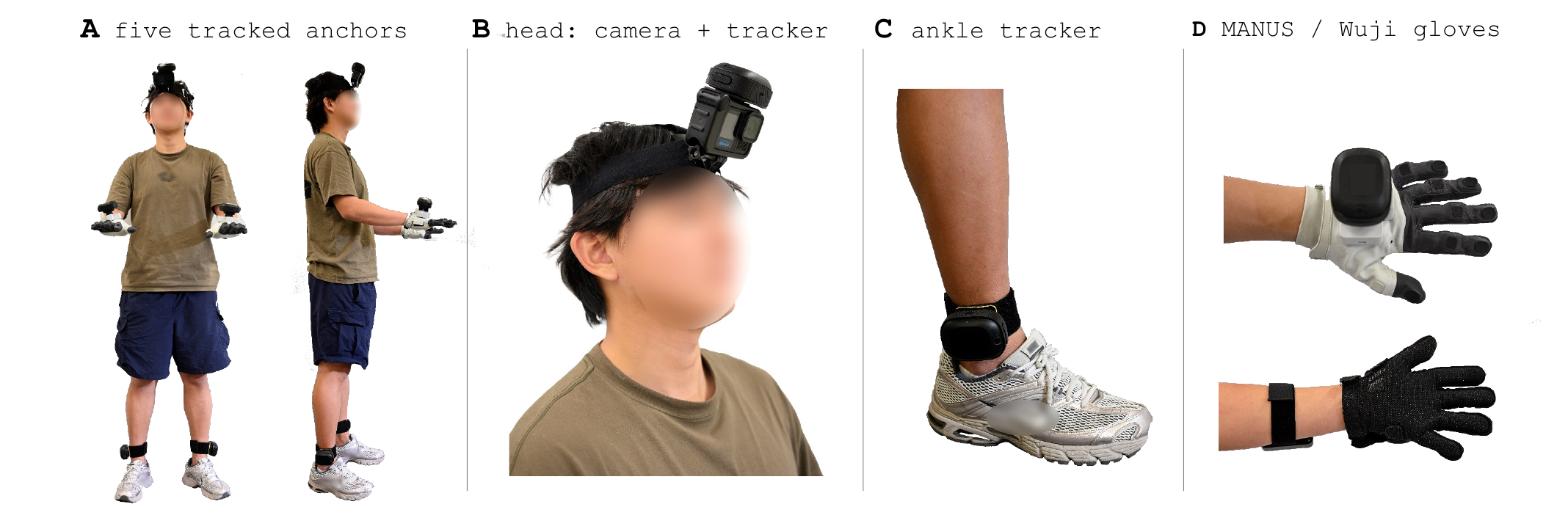}
\caption{Overview of the \name capture rig. Five body trackers are mounted
on the head, wrists, and ankles. MANUS gloves
capture finger articulation, while Wuji gloves additionally provide tactile
sensing.}
\label{fig:system}
\end{figure}}

Extending \cref{sec:system}, the capture system comprises a head-mounted camera, five wearable body trackers, and articulation gloves; the Wuji configuration additionally provides touch sensing. As in \cref{sec:system}, custom 3D-printed mounts rigidly couple the head tracker to the camera and attach the remaining trackers to the glove dorsum and the lateral ankles, with straps for added fixation (\cref{fig:system}). The trackers directly measure device poses, while anatomical anchors are obtained as joint frames through calibrated device-to-joint transformations. \cref{tab:sensor-spec} summarizes the acquisition settings and manufacturer-specified characteristics of the articulation gloves. Each sensing stream preserves its native timestamps for temporal alignment, and subsequent resampling only interpolates existing measurements without introducing additional observations. The calibration parameters are reused as long as the corresponding mounts remain fixed and are recomputed after remounting.

\begin{table}[H]
\centering\footnotesize
\caption{Capture hardware and acquisition settings. Glove rates are manufacturer
specifications; per-recording timing follows native timestamps.}
\label{tab:sensor-spec}
\begin{tabularx}{\linewidth}{@{}>{\raggedright\arraybackslash}p{0.18\linewidth}>{\raggedright\arraybackslash}p{0.33\linewidth}>{\raggedright\arraybackslash}X@{}}
\toprule
Stream & Hardware & Acquisition / representation\\
\midrule
Ego RGB & GoPro HERO13 Black with Media Mod & 30\,Hz; $1920\times970$ pixels; intrinsics below\\
Body tracking & Five VIVE Ultimate Trackers; head, wrists, ankles & 60\,Hz; room-fixed frame with its origin at a designated reference location\\
Finger articulation & MANUS Metagloves Pro (no haptic feedback) or Wuji gloves & Five 6-DoF fingertip EMF sensors per glove; 120\,Hz nominal rate\\
Regional touch & Wuji tactile configuration & $24\times31$ matrix; 744 taxels per hand; 526 retained by the evaluation health mask; 120\,Hz nominal rate\\
\bottomrule
\end{tabularx}
\end{table}

The camera operates at a resolution of $1920\times970$ pixels, with intrinsic matrix:
\[
K=\begin{bmatrix}
1012.0 & 0 & 960.0\\
0 & 1016.0 & 430.0\\
0 & 0 & 1
\end{bmatrix},
\]
where the focal lengths and principal-point coordinates are expressed in pixels. For body--hand fusion, glove joint positions are transformed into the wrist-local coordinate frame. The tactile sensing matrix follows row-major indexing and is mapped to 17 anatomical regions: two regions on the thumb, three on each remaining finger, and three on the palm. The contact extraction procedure used for quantitative evaluation is specified in \cref{app:glove-evaluation}, while the corresponding export interface is described in \cref{app:hand-fusion}. \cref{fig:system} illustrates the five body trackers, the rigid head mount coupling the camera to the head tracker, and the two interchangeable glove configurations.

\FloatBarrier

\section{Data Format}
\label[appendix]{app:schema}

The release is organized according to the LeRobotDataset v3.0 specification~\citep{cadenelerobot}, with all streams aligned to a common 30\,Hz temporal grid. Each row corresponds to one episode timestamp, and associates the corresponding visual observation, motion, tactile, and available annotation streams. The canonical fields \texttt{episode\_index}, \texttt{frame\_index}, \texttt{index}, \texttt{timestamp}, and \texttt{task\_index} denote the episode, local frame, global row, episode-relative time, and task, respectively. Timestamps are episode-relative and reset to zero at the beginning of each episode. Frame-level numerical data and textual annotations are stored in Parquet, whereas egocentric frames are encoded as MP4 and retrieved via episode metadata and temporal offsets. A shard may contain multiple episodes. The release layout is as follows:

\begin{quote}\footnotesize\ttfamily\raggedright
meta/info.json; meta/stats.json; meta/tasks.parquet\\
meta/episodes/chunk-000/file-000.parquet\\
data/chunk-000/file-000.parquet\\
videos/observation.images.ego/chunk-000/file-000.mp4
\end{quote}

\texttt{meta/info.json} specifies \texttt{fps=30}, feature shapes and dtypes, and shard paths; \texttt{meta/stats.json} stores feature statistics; and \texttt{meta/tasks.parquet} provides a mapping from task indices to task descriptions. Episode metadata specify row ranges and video offsets, together with anonymized session, split, and calibration-profile identifiers. \cref{tab:release-schema} lists the per-row fields.

\begin{table}[H]
\centering\footnotesize
\caption{Contents associated with each release row. Dataset-specific
features extend LeRobot's standard indexing fields.}
\label{tab:release-schema}
\begin{tabularx}{\linewidth}{@{}>{\raggedright\arraybackslash}p{0.21\linewidth}>{\raggedright\arraybackslash}p{0.22\linewidth}>{\raggedright\arraybackslash}X@{}}
\toprule
Field group & Per-row shape / type & Contents and conventions\\
\midrule
Index and time & Scalars & Episode, frame, global row, task index, and timestamp in seconds\\
Ego RGB & Linked video frame & \texttt{observation.images.ego}; frame matched to the row timestamp\\
Tracker poses & $5\times3$; $5\times4$ & Position in meters; quaternion order $(x,y,z,w)$; per-device validity\\
Body motion & $3+3+63$ & World pelvis position $u_t$ (m), root orientation, and 21 local body-joint rotations (axis-angle, rad)\\
Hand motion & $2\times45$ & Left/right SMPL-X finger rotations in local axis-angle form\\
Tactile readings & $2\times24\times31$ & Normalized taxel readings with invalid cells masked; tactile subset only\\
Regional contact & $2\times17$ & Binary region states over each 30\,Hz interval, with separate validity\\
Step annotations & Integers, text, times & Step ID, description, original interval bounds, skill, object roles, and observed outcome\\
Validity and provenance & Arrays and masks & Stream support, source-time mapping, interpolation flags, diagnostics, and review status; references where available\\
\bottomrule
\end{tabularx}
\end{table}

Step annotations will accompany every row in their annotated interval through
additional \texttt{annotation.*} fields; their original boundaries will be
retained in seconds. Episode-level task descriptions and step-level text remain
distinct. Fixed shape, scale, and mount parameters $\Phi$ will be referenced
by calibration-profile ID instead of repeated in every row. Measurements,
reconstructions, and reference labels will retain separate provenance;
unavailable streams will be marked by validity masks rather than encoded as
measured zeros.

Body and tracker positions use a metric, right-handed, $+Y$-up world frame.
Tracker quaternions use $(x,y,z,w)$ order and active device-to-world rotations
on column vectors. SMPL-X rotations are local axis-angle vectors in radians;
the root orientation is world-relative. The release translation field stores
the world pelvis position $u_t$ defined in \cref{sec:body-hand-reconstruction},
rather than the SMPL-X library's additive translation parameter.
The conversion is specified in \cref{app:body-coordinates}.

\textbf{Temporal alignment and validity.}
Native host timestamps are retained and converted to episode-relative seconds
using their recorded time origins. The 30\,Hz grid uses linear interpolation
for positions, spherical linear interpolation (SLERP)~\citep{shoemake1985slerp} for rotations, and
nearest recorded timestamps for RGB. Interpolated or held values retain
source-time and validity information; they are not new sensor measurements.
For contact, each output interval retains any valid native-rate detection
within that interval; a bin without valid samples remains unavailable
(\cref{app:hand-fusion}). Original step boundaries are stored in seconds,
with per-row annotations assigned by interval membership.

\FloatBarrier

\section{Whole-Body Reconstruction from Sparse Measurement}
\label[appendix]{app:reconstruction}

This section supplements $\mathrm{C}^3$ in \cref{sec:pipeline}, following the
same calibration, completion, and constraint stages. We retain its profile
$\Phi$, calibration variables $\Theta$, motion state $x_t$, and losses in
\cref{eq:calibration-forward,eq:calibration-objective,eq:body-objective}.
All distances inside the optimization are in meters. Below,
$\operatorname{MS}$ denotes the mean of squared scalar entries,
$[\,\cdot\,]_+=\max(\cdot,0)$, and $\|\cdot\|_2$ and $\|\cdot\|_F$ are Euclidean and
Frobenius norms.

\subsection{Protocol-guided calibration}
\label[appendix]{app:calibration-method}

\begin{figure}[!tp]
  \centering
  \includegraphics[width=0.99\linewidth,height=0.80\textheight,keepaspectratio]{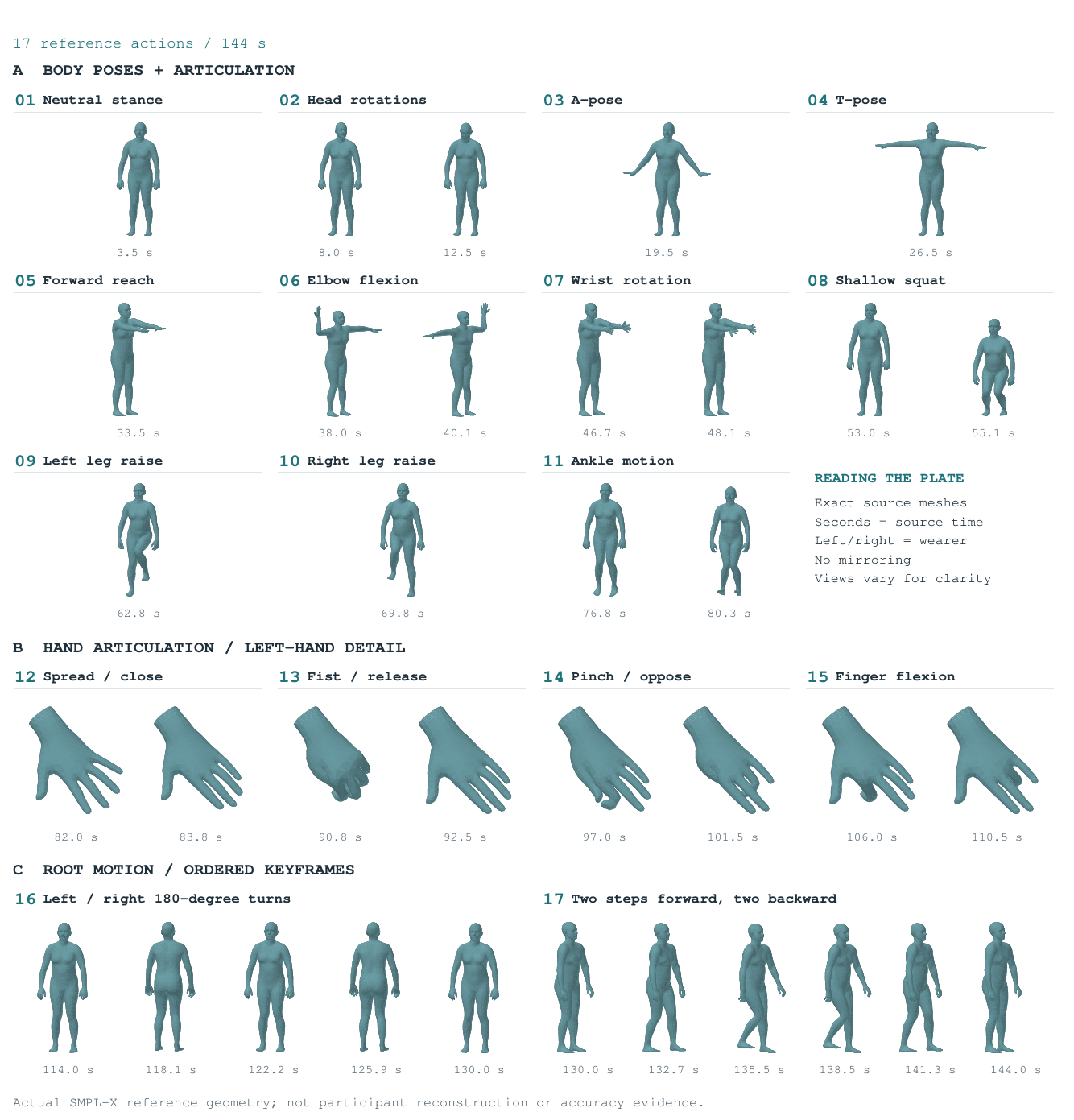}
  \caption{\textbf{Protocol-guided body--sensor calibration sequence.}
  Model-generated reference motions across 17 stages (144\,s): (A) static holds
  and body articulation; (B) finger actions; (C) turning and stepping.
  These illustrate the protocol, not measured participant motion.}
  \label{fig:calibration-protocol-real}
\end{figure}

\textbf{Reference protocol.}
\cref{fig:calibration-protocol-real} shows the 17 ordered stages used by
\cref{sec:lightweight-calibration}, named as in the figure: (A)~body poses
and articulation, (01)~neutral stance, (02)~head rotations, (03)~A-pose,
(04)~T-pose, (05)~forward reach, (06)~elbow flexion, (07)~wrist rotation,
(08)~shallow squat, (09)~left leg raise, (10)~right leg raise, and
(11)~ankle motion; (B)~hand articulation, (12)~spread/close,
(13)~fist/release, (14)~pinch/oppose, and (15)~finger flexion; and
(C)~root motion, (16)~left/right $180^\circ$ turns and (17)~two steps forward,
two backward. Stages (01)--(11) cover the body actions listed in the main
text, where neutral stance corresponds to standing and forward reach to
forward arm extension. A VPoser decoder
\citep{pavlakos2019smplx} fits body landmark targets, followed by local
elbow, wrist, head, and ankle corrections. Finger targets use the
MANO-compatible hand model \citep{romero2017mano}. Turning and stepping
add prescribed root motion and alternating support. The cached library
provides the reference joint positions $J_{n,i}(\beta)$ and rotations
$G_{n,i}$ already defined in \cref{eq:calibration-forward}; joint positions
are reevaluated at the participant's shape during fitting.

\textbf{Trajectory-based alignment and sample selection.}
Following \cref{sec:lightweight-calibration}, each sample $n$ comes from
recording $r(n)$ and is paired with reference action $k_n$ at phase
$\alpha_n$. The ordered protocol identifies the action, while motion
patterns and reviewed boundaries locate its interval in the recording.
Within that action, we match scale-normalized pairwise tracker distances
and relative orientation changes using monotonic dynamic time warping
(DTW)~\citep{sakoe1978dtw}. Reviewed events in a reference recording link characteristic poses
to library phases; the matching path transfers these correspondences to
recording $r(n)$ while accommodating different speeds and pauses.
Thus $\alpha_n$ is the matched reference phase, rather than a fixed fraction
of elapsed recording time. The resulting observed poses $p_{n,i},R_{n,i}$
are paired with the reference joint poses for $(k_n,\alpha_n)$ used in
\cref{eq:calibration-forward}.

For static holds, we aggregate stable observations after removing
transitions; for dynamic actions, we aggregate valid observations within
$\pm0.11$\,s of each matched instant. Positions use coordinatewise medians
and orientations use rotation averaging~\citep{hartley2013rotation}. Invalid or geometrically implausible
tracker streams are masked by $m_{n,i}$; holds and dynamic samples receive
weights $w_n=1$ and $0.1$, respectively. Finger, turning, and stepping
stages are retained for motion checks but excluded from the body-profile
fit. Matched phases, timestamps, and validity are recorded with each sample.

\textbf{Step 1: jointly fit body geometry and attachment positions.}
With the action correspondences fixed, we solve
\cref{eq:calibration-objective}, whose three terms are expanded in
\cref{eq:calibration-position-residual,eq:calibration-anthropometric,eq:calibration-geometry-regularizer},
for the main-text variables $\Theta$:
body shape and scale, the five attachment translations $a_i$, and each
recording's alignment $(Q_r,u_r)$. Each $a_i$ is one anchor-local vector
shared across all matched poses; reference articulation remains fixed.
At each iteration, we reevaluate the SMPL-X joints at the current shape,
predict tracker positions using \cref{eq:calibration-forward}, and
backpropagate the calibration loss through these operations. Adam then
updates all components of $\Theta$ jointly. The attachment rotations $A_i$
are computed separately in Step 2.

We initialize shape and scale as in \cref{sec:lightweight-calibration},
set $a_i=0$, and initialize each $Q_r$ from the horizontal left--right
wrist direction in the T-pose. The initial $u_r$ is the coordinatewise
median of observed-minus-predicted tracker positions for that recording.
We optimize $\log s$ to keep scale positive. Only the first five shape
coefficients are updated; the remaining five retain the visual estimate.

\emph{Robust tracker-position loss ($\mathcal L_{\mathrm{robust\mbox{-}pos}}$).}
A calibrated body and its fixed attachments should explain the measured
tracker locations across the matched actions. Tracking spikes and imperfect
execution of the reference poses can nevertheless produce large residuals.
We use a Huber penalty~\citep{huber1964robust} with threshold
$\tau=0.025$\,m: small errors receive a quadratic penalty for precise
fitting, while large errors receive a linear penalty so that a few poor
observations do not dominate the calibration.
For squats, the pelvis also moves while the feet remain planted. A shared
translation $\delta_n$, computed as the mean observed-minus-predicted
position of valid ankle trackers, accounts for this displacement without
changing the mounts. It is recomputed each iteration and is zero for other
actions or when no ankle is valid. With
$e_{n,i}=\widehat p_{n,i}+\delta_n-p_{n,i}$ and predictions from
\cref{eq:calibration-forward}, the weighted position loss is:
\begin{equation}
 \mathcal L_{\mathrm{robust\mbox{-}pos}}=
 \frac{\sum_{n,i,c}w_nm_{n,i}\rho_{0.025}(e_{n,i,c})}
 {3\max(1,\sum_{n,i}w_nm_{n,i})}.
 \label{eq:calibration-position-residual}
\end{equation}
Here $c$ indexes Cartesian coordinates, and the denominator averages over
valid weighted observations and their three coordinates. Distances are in
meters; $\rho_\tau(e)=e^2/2$ for $|e|\leq\tau$ and
$\tau(|e|-\tau/2)$ otherwise. Invalid entries are removed before arithmetic.

\emph{Measured body-size loss ($\mathcal L_{\mathrm{anth}}$).}
Tracker agreement alone cannot reliably distinguish a longer limb from a
larger mounting offset. We therefore require the fitted body to agree with
the participant's measured height and arm span. Using the main-text model
height $H(\beta)$, the unscaled T-pose arm span $W(\beta)$, and measured
values $h^\star,w^\star$, this loss is:
\begin{equation}
 \mathcal L_{\mathrm{anth}}=
 (sH(\beta)-h^\star)^2+(sW(\beta)-w^\star)^2.
 \label{eq:calibration-anthropometric}
\end{equation}
Both measurements are in meters. The weights in
\cref{eq:calibration-objective} are $\lambda_{\mathrm{obs}}=420$ and
$\lambda_{\mathrm{anth}}=1800$.

\emph{Geometry regularization ($\mathcal R_{\mathrm{geom}}$).}
Even with measured body size, several body--mount configurations may explain
the sparse observations. We favor the visually estimated proportions,
plausible attachment locations, and small scale and yaw corrections.
Using the mean-square operator $\operatorname{MS}$ defined above, the
implemented regularizer is:
\begin{equation}
\begin{aligned}
\mathcal R_{\mathrm{geom}}={}&
0.012\operatorname{MS}(\beta)
+0.18\operatorname{MS}(\beta_{1:5}-\beta^{\mathrm{vis}}_{1:5})\\
&+\frac{280}{5}\sum_{i=1}^{5}[\|a_i\|_2-\ell_i]_+^2\\
&+3\!\left[(\|a_2\|_2-\|a_3\|_2)^2
           +(\|a_4\|_2-\|a_5\|_2)^2\right]\\
&+12\operatorname{MS}\!\left(
 \{G_{\mathrm{neutral},i}a_i-d_i\}_{i\in\{4,5\}}\right)\\
&+0.18(\log s-\log s_0)^2
 +0.015\operatorname{MS}(\{\psi_r\}_r).
\end{aligned}
\label{eq:calibration-geometry-regularizer}
\end{equation}
Here $\ell=(0.18,0.15,0.15,0.18,0.18)$\,m gives soft mount-length
limits in head, left/right wrist, left/right ankle order. The bilateral
terms compare mount lengths, allowing mirrored local axes.
$G_{\mathrm{neutral},i}$ is anchor $i$'s reference orientation in the
neutral-stance stage~(01); it expresses the ankle offset in neutral body axes before
comparison with $d_4=(0.075,0.070,0)$ and
$d_5=(-0.075,0.070,0)$\,m. The corresponding mean square averages all six
ankle-offset coordinates. Finally, $\psi_r$ is the yaw angle of $Q_r$.
These soft penalties limit body--mount compensation while allowing the
tracker observations and measured body dimensions to determine the fit.

Adam~\citep{kingma2014adam} runs for 1,100 iterations, with learning rates
0.008 for log-scale/yaw, 0.018 for shape/recording translation, and 0.012
for attachment translation. After each update, we clamp scale to
$[0.72,1.15]$, optimized shape coefficients to $[-3,3]$, and yaw to
$[-\pi,\pi]$. This returns $\Theta^{\mathrm{fit}}$, including the shared
$a_i$; no per-frame attachment is fitted.

\textbf{Step 2: recover attachment rotations by robust averaging.}
With the fitted $Q_r^{\mathrm{fit}}$ fixed, the orientation relation in
\cref{eq:calibration-forward} gives one candidate device-to-anchor rotation
$\widetilde A_{n,i}$ per valid observation. Let $\mathcal I_i$ contain the retained
samples for tracker $i$: static holds for wrists, and both static and
selected dynamic samples for the head and ankles. Compute the mean
candidate matrix $M_i$, take its singular value decomposition (SVD), and
project it to a proper rotation:
\begin{equation}
\begin{aligned}
 \widetilde A_{n,i}&=(Q_{r(n)}^{\mathrm{fit}}G_{n,i})^\top R_{n,i},\\
 M_i&=\frac{1}{|\mathcal I_i|}\sum_{n\in\mathcal I_i}\widetilde A_{n,i}
      =U_i\Sigma_iV_i^\top,\\
 A_i&=U_i\operatorname{diag}\!\left(1,1,\det(U_iV_i^\top)\right)V_i^\top.
\end{aligned}
\label{eq:calibration-rotation-recovery}
\end{equation}
The determinant correction excludes reflections. For a fixed retained set,
this gives the rotation minimizing the summed squared Frobenius distances
to its candidates; since $A_i$ is the right factor of $\widehat R_{n,i}$ in
\cref{eq:calibration-forward} and orthogonal factors preserve the Frobenius
norm, it equivalently minimizes
$\sum_{n\in\mathcal I_i}\|\widehat R_{n,i}-R_{n,i}\|_F^2$, i.e., it matches
the predicted orientations $\widehat R_{n,i}$ to the measured $R_{n,i}$ as in
\cref{sec:lightweight-calibration}. It requires no gradient optimization.

We make the estimate robust with up to three rejection rounds. After each
average, compute each candidate's angular residual as the rotation angle
of $A_i^\top\widetilde A_{n,i}$. Reject residuals above the larger of $12^\circ$ and
the median plus 2.5 times the median absolute deviation (floored at
$1^\circ$), then recompute the average. Stop if the set is unchanged or a
rejection would leave fewer than four samples. An empty initial set uses
an identity fallback marked as low confidence. The resulting five $A_i$,
together with the fitted shape, scale, and $a_i$, complete the reusable
profile $\Phi$. \Cref{eq:calibration-rotation-recovery} with this rejection
rule is the explicit implementation of $f_{\mathrm{rot}}$ in
\cref{sec:lightweight-calibration}.

\subsection{Prior-guided motion completion}
\label[appendix]{app:proposal-training}
\label[appendix]{app:body-coordinates}

\textbf{Shared state and coordinate conventions.}
We use $x_t=(\omega_t,\theta_t,u_t)$ and the calibrated anchor frames
$\widetilde p_{t,i},\widetilde R_{t,i}$ from
\cref{sec:body-hand-reconstruction}. Here $u_t$ is the world pelvis
position. The SMPL-X library's additive translation subtracts the scaled
zero-translation pelvis location from $u_t$; conversion back to $u_t$ is
applied on export. This is an implementation coordinate conversion, not an
additional motion variable. The five anchors are SMPL-X joints
$(15,20,21,7,8)$ in head, left/right wrist, left/right ankle order.
For task-time fitting, \cref{eq:calibration-forward} uses the rotation
represented by $\omega_t$ in place of $Q_r$, joint geometry evaluated from
$\theta_t$ in place of the reference pose, and $u_t$ in place of $u_r$.
No second yaw is applied, and metric attachment offsets $a_i$ are never
scaled again.

\textbf{Input representation and model.}
The feature vector $\mathbf z_t$ from the main text has 90 entries:
for each anchor, six rotation coordinates, six relative-rotation coordinates,
three positions, and three position increments. The six-dimensional
rotation representation uses the first two matrix columns
\citep{zhou2019continuity}; relative rotations are previous-rotation
transpose times current rotation. Channels are packed by feature type for
head/wrists (54), then ankles (36). Initial rotation increments are identity
and position increments zero; increments are not divided by time.

The AvatarPoser-based~\citep{jiang2022avatarposer}
transformer~\citep{vaswani2017attention} prior $f_{\mathrm{prior}}$ has width
256, three encoder layers, and eight heads. The last token of an $L=40$
frame trailing window predicts 22 rotations (root and 21 body joints) plus
two auxiliary foot-contact logits. Initial windows are shorter. Rotation
outputs are orthonormalized, interpolated back to tracker timestamps, and
converted to capture coordinates. The head-alignment expression for
$u_t^0$ in \cref{sec:body-hand-reconstruction} then gives $\mathcal X^0$;
it is not a second fitting stage.

\textbf{Training data and optimization.}
Compatible public AvatarPoser weights initialize the network; the added
36 ankle input channels are zero-initialized. The pool combines 2,700
AMASS-derived sequences \citep{mahmood2019amass} from HDM05, CMU, and BMLrub,
VPoser synthetic motion, 20 sequences from the ten B01--B10 guides at
two speeds/durations, and teacher-fitted real recordings, including
repetitions 1--2 of the full-body evaluation actions
(\cref{app:evaluation-data}). Synthetic tracker inputs use participant
shape, scale, and mounts. VPoser latent random walks last 240 frames at
60\,Hz, with 2\,mm Gaussian position noise and 1.5\% simulated dropout.
Wrist-twist augmentation duplicates 30\% of synthetic sequences with a
$\pm25^\circ$ twist in one wrist observation and its target.

Teacher fits use 300 Adam iterations per chunk, learning rate 0.01, and
$(\lambda_p,\lambda_R,\lambda_{\mathrm{prior}})=(10000,60,0.5)$, retaining the
other active constraints. They provide task-domain supervision rather than
independent ground truth. Each real recording uses its own calibration and
up to 2,000 uniformly spaced 40-frame windows. Training minimizes mean
SmoothL1 loss (unit threshold) on 132 rotation coordinates plus 0.25 times
mean binary cross-entropy on the two foot logits. Natural-sequence foot
targets are zero; this auxiliary head is not used as support evidence in
refinement. AdamW \citep{loshchilov2017adamw} runs for three epochs, with
learning rate and weight decay $10^{-4}$, batch size 32, and gradient-norm
clipping at 1. The 60\,Hz network timeline is internal; the release remains
30\,Hz. Evaluation repetition 3 is held out as specified in
\cref{app:evaluation-data}. Video and glove signals are not network inputs.

\subsection{Constraint layer 1: skeletal motion}
\label[appendix]{app:skeletal-losses}

The weighted sum $\mathcal L_{\mathrm{skel}}$ is defined in
\cref{sec:layered-constraints,eq:body-objective}; we expand its five
component losses here.
Within one of the temporal chunks of \cref{sec:layered-constraints}, with
$B$ frames, $t=1,\ldots,B$ is a local frame index and
$i=1,\ldots,5$. Let $m_{t,i}\in\{0,1\}$ indicate a valid tracker
observation, and $N_{\mathrm{obs}}=\max(1,\sum_{t=1}^{B}\sum_{i=1}^{5}m_{t,i})$. The
implemented measurement losses are coordinate-normalized squared errors:
\begin{equation}
  \mathcal L_p=\frac{1}{3N_{\mathrm{obs}}}\sum_{t,i}m_{t,i}
  \|\widehat p_{t,i}-p_{t,i}\|_2^2,\qquad
  \mathcal L_R=\frac{1}{9N_{\mathrm{obs}}}\sum_{t,i}m_{t,i}
  \|\widehat R_{t,i}-R_{t,i}\|_F^2.
  \label{eq:tracker-losses}
\end{equation}
Predictions come from \cref{eq:calibration-forward} and are compared with raw
device observations $p_{t,i},R_{t,i}$, as specified in the main text.
The inferred anchor frames in \cref{sec:body-hand-reconstruction} are used for completion. Dividing by 3 and 9 averages scalar
components; it does not remove the difference in physical units.
The rotational objective is chordal matrix distance, not the geodesic angle
used by diagnostic reporting. For the chunk of $\mathcal X^0$, the regularizers are:
\begin{align}
  \mathcal L_{\mathrm{prior}}&=
  \operatorname{MS}(\theta-\theta^0)
  +0.2\operatorname{MS}(\omega-\omega^0)
  +0.2\operatorname{MS}(u-u^0),\label{eq:proposal-prior}\\
  \mathcal L_{\mathrm{time}}&=
  \operatorname{MS}(\{\theta_t-\theta_{t-1}\}_{t=2}^{B}).
  \label{eq:body-temporal}
\end{align}
Here $\theta,\omega,u$ stack the chunk's main-text states and superscript
$0$ denotes the proposal. With fixed shape, the library's additive-translation
offset cancels in $u-u^0$. The temporal term is zero for $B=1$.
Rotation regularizers use axis-angle differences; none of these differences
is normalized by elapsed time. The active
configuration neither enforces temporal coupling across chunk boundaries nor
uses overlapping chunks.

Let $\kappa_{t,l}$ and $\epsilon_{t,l}$ be geometric knee and elbow flexion
angles, with zero at straight extension, and $\eta_t$ the pelvis-to-neck
angle from the upward axis. The active soft-angle loss is:
\begin{equation}
\begin{split}
  \mathcal L_{\mathrm{joint}}={}&
  \frac{1}{2B}\sum_{t,l}[\kappa_{t,l}-125\pi/180]_+^2
  +\frac{1}{2B}\sum_{t,l}[\epsilon_{t,l}-140\pi/180]_+^2\\
  &+\frac{1}{B}\sum_t[\eta_t-15\pi/180]_+^2.
\end{split}
\label{eq:soft-angle-limits}
\end{equation}
Here $l$ indexes the two sides. The torso threshold is an engineering bias,
not an anatomical limit; genuine deep bends may be penalized. This term belongs to $\mathcal L_{\mathrm{skel}}$ in
\cref{eq:body-objective}, even though its angles are evaluated from
posed joint geometry.

\subsection{Constraint layer 2: surface clearance}
\label[appendix]{app:surface-loss}

We evaluate the posed SMPL-X surface already used in the main text's
surface and ground layers. Let $v_{t,j}\in\mathbb R^3$ denote surface
vertex $j$ in the tracking world frame, obtained from $x_t$ with fixed
$(\beta,s)$. Vertices are model outputs, not additional measurements or
optimization variables.
The six sampled parts are the torso, head, left arm, right arm, left leg, and
right leg. Their 30 ordered distinct-part pairs form $\mathcal P$. Skinning
weights assign a vertex to its strongest part when that part's weight exceeds
0.8; two boundary-erosion steps exclude transition regions, and farthest-point
sampling~\citep{qi2017pointnetpp} selects 256 interior vertices per part on the fixed neutral mesh.
Let $\mathcal S_b$ contain the sampled vertex indices of part $b$; these are
the ``points sampled from distinct body parts'' in \cref{sec:layered-constraints}.
For $j\in\mathcal S_b$, let $j'$ be its nearest sampled vertex
on target part $b'$, and let $\mathbf N_{t,j'}$ be that vertex's area-weighted
unit surface normal. The signed clearance $d_{t,j}^{b,b'}$ is positive outside
the target surface and negative once it is crossed; the 0.10\,m gate restricts
the penalty to nearby points, and TopK aggregates the largest violations per
part pair. The adopted oriented penalty is:
\begin{equation}
\begin{split}
 d_{t,j}^{b,b'}&=\langle v_{t,j}-v_{t,j'},\mathbf N_{t,j'}\rangle,\\
 q_{t,j}^{b,b'}&=\mathbf1\{\|v_{t,j}-v_{t,j'}\|_2<0.10\}
                  [0.018-d_{t,j}^{b,b'}]_+^2,\\
 \mathcal L_{\mathrm{surf}}&=\frac{1}{B|\mathcal P|K}
  \sum_{t=1}^{B}\sum_{(b,b')\in\mathcal P}
  \sum_{j\in\operatorname{TopK}(q_t^{b,b'})}q_{t,j}^{b,b'},\qquad K=8.
\end{split}
\label{eq:surface-penalty}
\end{equation}
Distances are in meters; TopK selects the eight largest penalties, including
zeros when fewer than eight are active. Nearest indices, normals, and the
0.10\,m support gate are recomputed each iteration without gradients; gradients
flow through both selected vertex positions. The 0.018\,m clearance discourages
local penetration, but the gate can omit deep intersections and this local
proxy is not an exact signed distance field or a collision-free guarantee.

\subsection{Constraint layer 3: body-ground consistency}
\label[appendix]{app:ground-loss}
The estimated ground plane of \cref{sec:layered-constraints} is the
horizontal plane at height $g$ in the $+Y$-up world frame. Summing over all
posed body vertices $v_{t,j}$ (\cref{app:surface-loss}), the floor term is:
\begin{equation}
  \mathcal L_{\mathrm{floor}}=
  \frac{1}{B}\sum_{t=1}^{B}\sum_j
  [g-v_{t,j}^{(y)}]_+^2.
  \label{eq:global-floor}
\end{equation}
The sum over vertices is intentionally not averaged, and the
superscript $(y)$ selects the vertical coordinate. Unless overridden, the
launcher sets $g$ to the fifth percentile of valid ankle-tracker heights
minus $0.12$\,m; this heuristic height estimate is used instead of fitting
a plane to scene geometry.

\subsection{Joint refinement}
\label[appendix]{app:body-solver}
The solver minimizes \cref{eq:body-objective} using the component losses
in \cref{eq:tracker-losses,eq:proposal-prior,eq:body-temporal,eq:soft-angle-limits,eq:surface-penalty,eq:global-floor},
with the calibrated profile fixed. In
$\mathcal L_{\mathrm{skel}}=\sum_k\lambda_k\mathcal L_k$, the weights are
$(\lambda_p,\lambda_R,\lambda_{\mathrm{prior}},\lambda_{\mathrm{time}},
\lambda_{\mathrm{joint}})=(3000,20,2,5,100)$; in \cref{eq:body-objective},
$\lambda_{\mathrm{surf}}=3000$ and $\lambda_{\mathrm{floor}}=50$. Contact, capsule, torso-hull, sole-only floor,
and sole-flatness weights are zero. The batch wrapper uses Adam~\citep{kingma2014adam} for 60
iterations per chunk with gradient-norm clipping at 2. Its initial learning
rate/chunk size is $(0.004,256)$; bounded fallbacks are $(0.002,256)$,
$(0.001,256)$, $(0.0005,256)$, $(0.001,128)$, and $(0.0005,128)$.
Each chunk starts from its states in $\mathcal X^0$. Every Adam iteration
evaluates all three layers and updates the state $x_t$ through the library translation conversion in
\cref{app:body-coordinates}; there are
no separate skeletal, surface, or ground optimization passes. Concatenating
the non-overlapping fitted chunks gives $\mathcal X^\star$. The argmin in
\cref{eq:body-objective} specifies the objective; the implementation
returns its finite-iteration numerical estimate, without a global-optimum
guarantee. Failed chunks use the bounded retry settings above.

\FloatBarrier

\section{Multimodal Integration and Annotation}
\label[appendix]{app:integration}

\subsection{Hand and touch fusion}
\label[appendix]{app:hand-fusion}

\textbf{Hand alignment and fitting.}
As in \cref{sec:annotation}, finger articulation is fitted in wrist-local
coordinates after body reconstruction. Glove positions are translated to
the measured wrist origin and rotated into its frame; wrist-local Wuji
inputs enter directly. Both layouts retain four landmarks per finger,
including the tip (20 non-wrist landmarks); the 25-node layout discards four
extra palm/metacarpal nodes. A sequence-fixed similarity aligns these points
to the calibrated model's flat hand, using the median of up to 12 frames
with the straightest fingers. Scale is the ratio of model/glove RMS radii
about the wrist; rotation uses SVD with reflection
correction~\citep{umeyama1991least}. This hand
scale is separate from body scale $s$.

The MANO-compatible SMPL-X hand \citep{romero2017mano,pavlakos2019smplx}
optimizes its 45 local axis-angle coordinates per side. The fitting objective is:
\begin{equation}
 \mathcal L_{\mathrm{hand}}=
 1000\mathcal L_{\mathrm{points}}+0.015\mathcal L_{\mathrm{directions}}
 +0.003\mathcal L_{\mathrm{flat}}+0.004\mathcal L_{\mathrm{continuity}}
 +0.02\mathcal L_{\mathrm{angle}}.
 \label{eq:hand-objective}
\end{equation}
The point term averages squared coordinate errors over the 20 wrist-local
landmarks in meters. The direction term averages one minus cosine similarity
along the 20 finger bones, with vector norms floored at $10^{-8}$.
The flat-hand term is mean squared pose deviation from the model's flat-hand
pose. Continuity averages squared adjacent-frame pose differences and adds
twice the mean squared difference from the preceding window's final pose;
the latter is omitted for the first window. The angle term averages squared
excess of each of the 15 joint-rotation magnitudes above 2.7 radians.

Left/right objectives \cref{eq:hand-objective} are summed. At most 480 frames are sampled over the
hands' overlapping time range and fitted in 90-frame windows with 60 Adam
iterations and learning rate 0.04. After accounting for host-time origins,
hand rotations are SLERPed onto the body timeline; endpoint holds outside
the fitted span retain coverage flags. Fusion replaces hand-pose arrays
while preserving the fitted body, root, shape, scale, and attachments.

\textbf{Tactile processing.}
Wearable tactile response weakens with use and depends on wearing tension
and channel health. Raw amplitude therefore cannot by itself distinguish
light contact from reduced sensitivity or missing data. The three interfaces
in \cref{tab:tactile-levels} retain native values (L0), baseline-relative
changes (L1), and 17-region contact with separate health/validity (L2).
Unresponsive taxels are excluded from regional support; insufficient support
makes a region unavailable, not negative. Baselines, response normalization,
and region mappings accompany each recording. These are contact indicators,
not calibrated force or pressure measurements.


Contact is temporally associated with motion, not used to constrain body IK. To preserve brief touches on the 30,Hz grid, each region's output is the logical OR of valid native-rate detections within the 1/30,s bin centered on the output timestamp. Bins without valid samples are masked. This release rule differs from the controlled press study's 30,Hz signal averaging, baselines, thresholds, and scoring masks (\cref{app:glove-evaluation}). The tactile L0--L2 interfaces are unrelated to the three body-constraint layers.

\begin{table}[H]
\centering\footnotesize
\caption{Tactile interfaces used by the release and evaluation. Binary contact
and channel validity are distinct; neither normalized amplitude nor a health
indicator is an absolute pressure measurement.}
\label{tab:tactile-levels}
\begin{tabularx}{\linewidth}{@{}lX@{}}
\toprule
Interface & Meaning\\
\midrule
L0: raw signal & Native taxel readings and validity, with range and any
sentinels defined by the device schema. Wearing-dependent offsets and gain remain.\\
L1: signal change & Changes relative to a no-contact baseline; still dependent
on sensor response and configuration.\\
L2: regional contact & Binary regional contact with validity/health recorded
separately. Regional mapping and threshold calibration are versioned.\\
\bottomrule
\end{tabularx}
\end{table}

\subsection{Expert-provided procedural context}
\label[appendix]{app:skill-structure}
\label[appendix]{app:text-skills}

The off-the-shelf annotation tools \citep{peng2026cortex} in
\cref{sec:annotation} receive video, task instructions, and expert-written
procedural context. \cref{tab:laboratory-skills} summarizes preparation,
reaction, monitoring, workup, extraction, concentration, and purification.
These descriptions supply terminology and likely procedural transitions;
observed activity determines which steps are annotated and in what order.

\begin{table}[H]
\centering\footnotesize
\caption{Example procedural knowledge supplied by laboratory Skills documents.}
\label{tab:laboratory-skills}
\begin{tabularx}{\linewidth}{@{}>{\raggedright\arraybackslash}p{0.21\linewidth}>{\raggedright\arraybackslash}X@{}}
\toprule
Experimental stage & Information provided to the annotation model\\
\midrule
Material preparation & Retrieve reagents, catalysts, containers, and consumables; bring them to the preparation bench and organize them by use. Preparation may include rinsing pipettes, taring a balance, weighing solids, and transferring material.\\
Reaction setup & Add materials to a reaction flask in the specified sequence, adjust heating and stirring, and monitor the apparatus. The document distinguishes reagent addition from device adjustment.\\
Reaction monitoring & For TLC, take a sample, spot a plate, and inspect it under UV illumination. For HPLC preparation, transfer and dilute a sample, mix it, filter it into a vial, and label the vial.\\
Quenching and workup & Move the reaction vessel to the workup area, handle receiving vessels, pour liquids, and inspect temperature or pH readouts as described in the procedure.\\
Extraction & Fill a separatory funnel, allow layers to settle, open and close the stopcock, collect layers in separate vessels, and repeat mixing and separation when required.\\
Rotary evaporation & Attach the flask, set the apparatus, operate vacuum/heating/rotation controls, monitor concentration, stop the apparatus, release vacuum, detach the flask, and weigh it.\\
Column chromatography & Mix the sample with silica, load and install the column, connect tubing, operate the apparatus, and collect fractions.\\
\bottomrule
\end{tabularx}
\end{table}

\subsection{Example text prompt}
\label[appendix]{app:skill-prompt}

The example below includes the procedural background itself, alongside the
annotation request. It is adapted from the laboratory document to illustrate
the input supplied to the model; it is not a verbatim archived run prompt.

\begin{quote}\small
\textbf{Task.} You are annotating a first-person recording of chemical
laboratory work. Divide the observed activity into meaningful procedural
steps and write a concise description of each step.

\textbf{Episode instruction.} \texttt{<task description for this recording>}\\
\textbf{Observations.} \texttt{<video or sampled frames with source timestamps>}

\textbf{Professional Skills: laboratory workflow context.}
The experiment may include the stages below. This information explains the
procedure and the operations you may see; the recording may cover only a
subset of them.

1. \emph{Preparation.} Reagents, catalysts, vessels, and consumables are
retrieved from storage, brought to the bench, and arranged by use. A balance
may be tared before solids are weighed and transferred; pipettes may be
rinsed before liquid handling.

2. \emph{Reaction setup.} Materials are added to a reaction flask. The
example procedure describes adding solvent, solid substrate, liquid
reagent, and catalyst in sequence, followed by adjustment of heating and
stirring and observation of the apparatus.

3. \emph{Monitoring.} TLC involves sampling, spotting a plate, and UV
inspection. HPLC sample preparation can involve liquid transfer, dilution,
mixing, filtration into a sample vial, and labeling.

4. \emph{Workup.} The reaction vessel may be moved from the stirring station
to the fume hood. Subsequent operations include handling the receiving
vessel, pouring, adjusting controls, and inspecting relevant readouts.

5. \emph{Extraction.} A mixture is transferred into a separatory funnel and
allowed to separate into layers. The stopcock is opened to drain a layer
into a receiver and closed to stop drainage; the other layer is collected
separately. Addition of extraction solvent, mixing, settling, and separation
may be repeated.

6. \emph{Rotary evaporation.} The flask is connected to the evaporator,
controls are adjusted, and concentration is monitored. Later operations
include stopping the apparatus, releasing vacuum, removing the flask,
and weighing it.

7. \emph{Column chromatography.} Sample and silica are mixed, material is
loaded into a column, the column and tubing are installed, and the apparatus
is operated to collect fractions.

\textbf{Annotation instructions.} Use this background to identify equipment
and interpret the stage of the experiment. Follow the video when deciding
which operations occurred and where their boundaries lie. Preserve visible
relocation, repetition, and correction. Describe the action and its target
concretely, such as ``attaches the flask to the rotary evaporator.'' Do not
insert steps solely because they appear in the procedure. Leave unreadable
quantities, uncertain identities, and unobserved outcomes unspecified.

\textbf{Output.} Return a time-ordered list of segments with
\texttt{start\_time}, \texttt{end\_time} (source-video seconds),
\texttt{skill}, \texttt{description}, \texttt{object\_roles}, and
\texttt{observed\_outcome}. Use a short operation label for \texttt{skill}
and \texttt{null} for an outcome that cannot be established from the video.
\end{quote}

\subsection{Human verification and experimental interpretation}
\label[appendix]{app:skill-review}

As stated in \cref{sec:annotation}, operators review the egocentric video,
reconstructed motion, available touch, and draft text together for temporal
alignment and semantic consistency. They check action and object
identification, omitted or extra steps, unsupported statements, and temporal
boundaries (\cref{app:review-protocol}). The ablation in \cref{tab:skills} adds or removes
the procedural document while retaining the same annotation model and other
inputs. It tests the benefit of professional background knowledge, with no
model retraining. Its automated draft assessment remains distinct from the
expert review in \cref{tab:language}.

\FloatBarrier

\section{Data Statistics}
\label[appendix]{app:stats}
\label[appendix]{app:stats-overview}

\begin{figure}[!tp]
\centering
\includegraphics[width=0.99\linewidth,height=0.80\textheight,keepaspectratio]{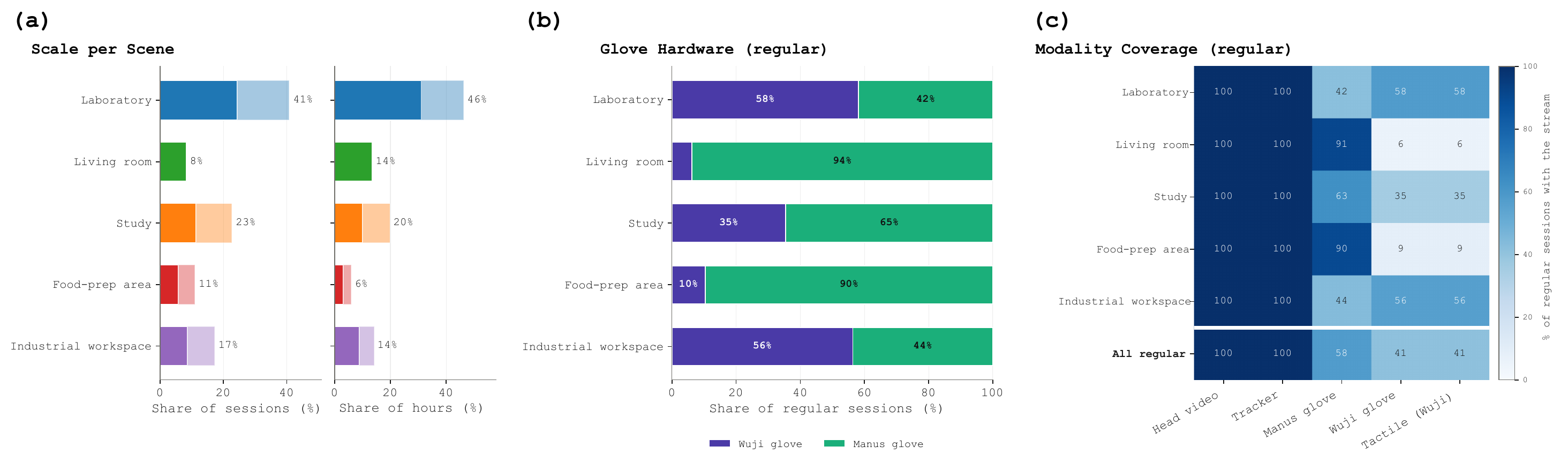}
\caption{Collection scale and coverage: (a) recording and duration shares by
environment; (b) glove configurations; (c) sensing-stream coverage.
Panels (b,c) use regular recordings.}
\label{fig:stats}
\end{figure}

\textbf{Collection scale and sensing coverage.}
\name contains 2{,}715 recordings totaling 100 hours across ten task
templates and five environments. Laboratory activities account for 41\% of
recordings and 46\% of duration; the remaining recordings span the living
room (8\%), study (23\%), food-preparation area (11\%), and industrial
workspace (17\%). \cref{fig:dataset-overview,fig:stats} distinguish regular
and recovery episodes and summarize their durations and sensing coverage.
Among regular recordings, head video and tracker streams are available
throughout; MANUS and Wuji glove streams cover 58\% and 41\%, respectively.
Measured touch accompanies the Wuji configuration and is available in
41\% of regular recordings. These rounded percentages describe stream
availability, not reconstruction accuracy. The analyses below use the
signals and health masks required by each diagnostic; their cohorts should
not be equated with the full collection.

\begin{figure}[!tp]
\centering
\includegraphics[width=0.99\linewidth,height=0.80\textheight,keepaspectratio]{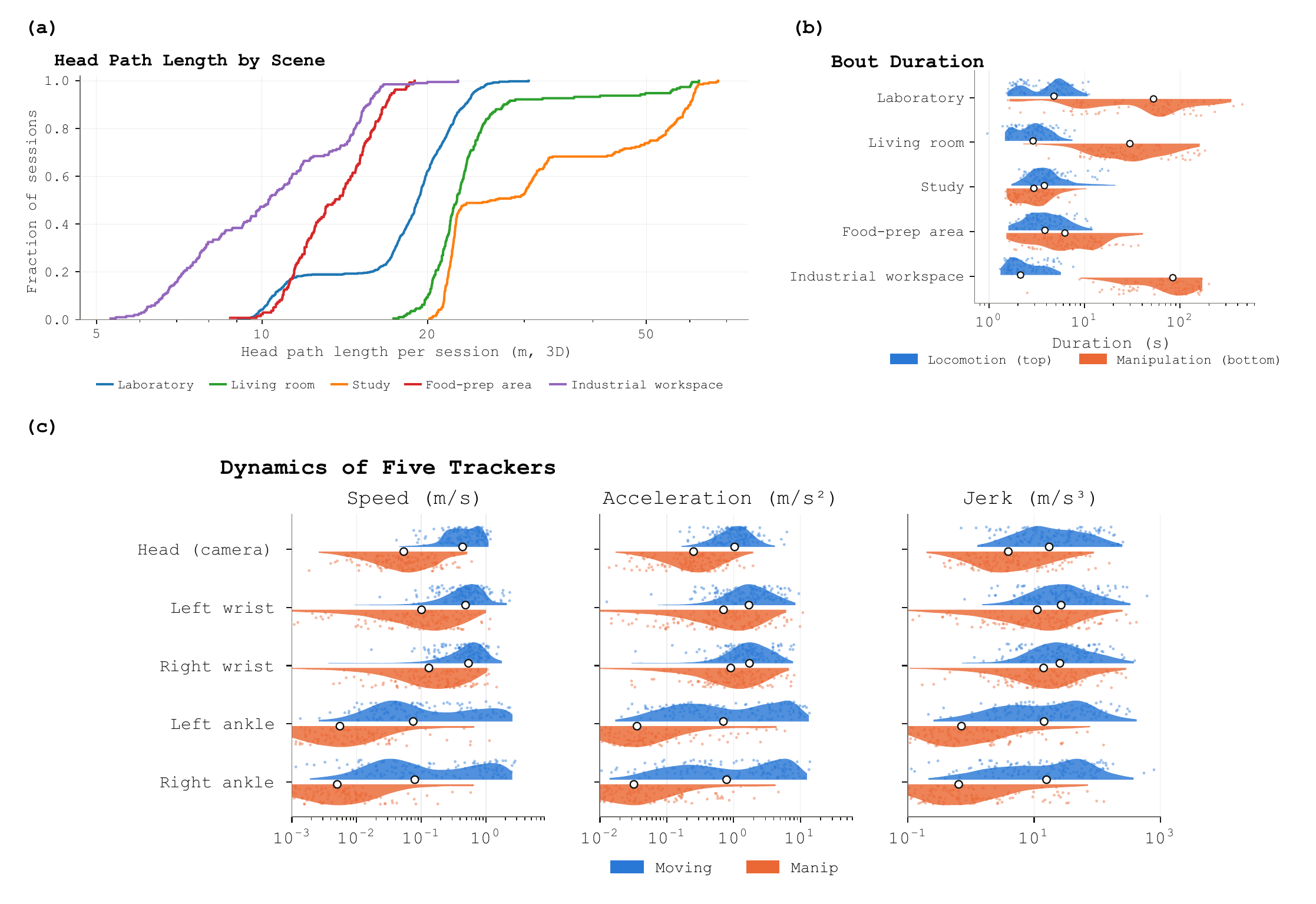}
\caption{Movement statistics: (a) 3D head paths; (b) locomotion and manipulation
bout durations; (c) five-tracker dynamics, with units indicated on the axes.}
\label{fig:motion-statistics}
\label{fig:behavior-statistics}
\end{figure}

\begin{figure}[!tp]
\centering
\includegraphics[width=0.99\linewidth]{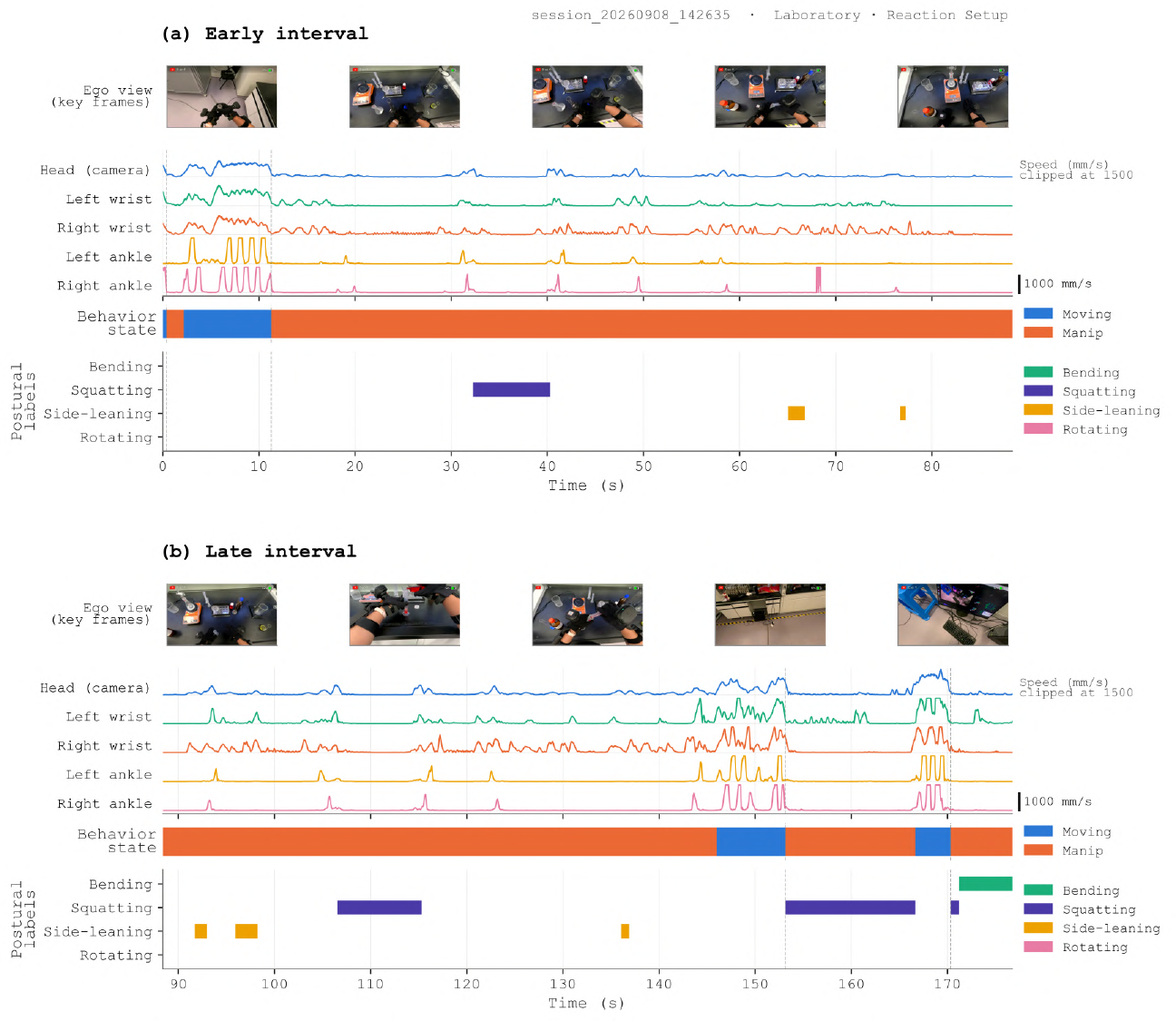}
\caption{A reaction-setup episode in two consecutive intervals. Rows align ego
frames, tracker speed, behavior, and posture; dashed lines mark detected
movement-to-manipulation transitions.}
\label{fig:episode-states}
\end{figure}

\textbf{Movement and task progression.}
\cref{fig:motion-statistics} summarizes 3D head-path length, locomotion and
manipulation bout duration, and the dynamics of all five trackers.
Short locomotion bouts alternate with sustained manipulation, particularly
in laboratory and industrial work. The study and living-room recordings
include longer head paths, while ankle dynamics distinguish locomotion from
stationary manipulation. Tracker speed, acceleration, and jerk describe
recorded behavior; reconstruction quality is evaluated separately in
\cref{tab:ablation}.
\cref{fig:episode-states} illustrates the temporal structure in a
reaction-setup episode, aligning ego frames with motion and detected posture
across two consecutive intervals. Displayed speed is capped at
1,500\,mm/s; the detected states are descriptive diagnostics.

\begin{figure}[!tp]
\centering
\includegraphics[width=0.80\linewidth,height=0.80\textheight,keepaspectratio]{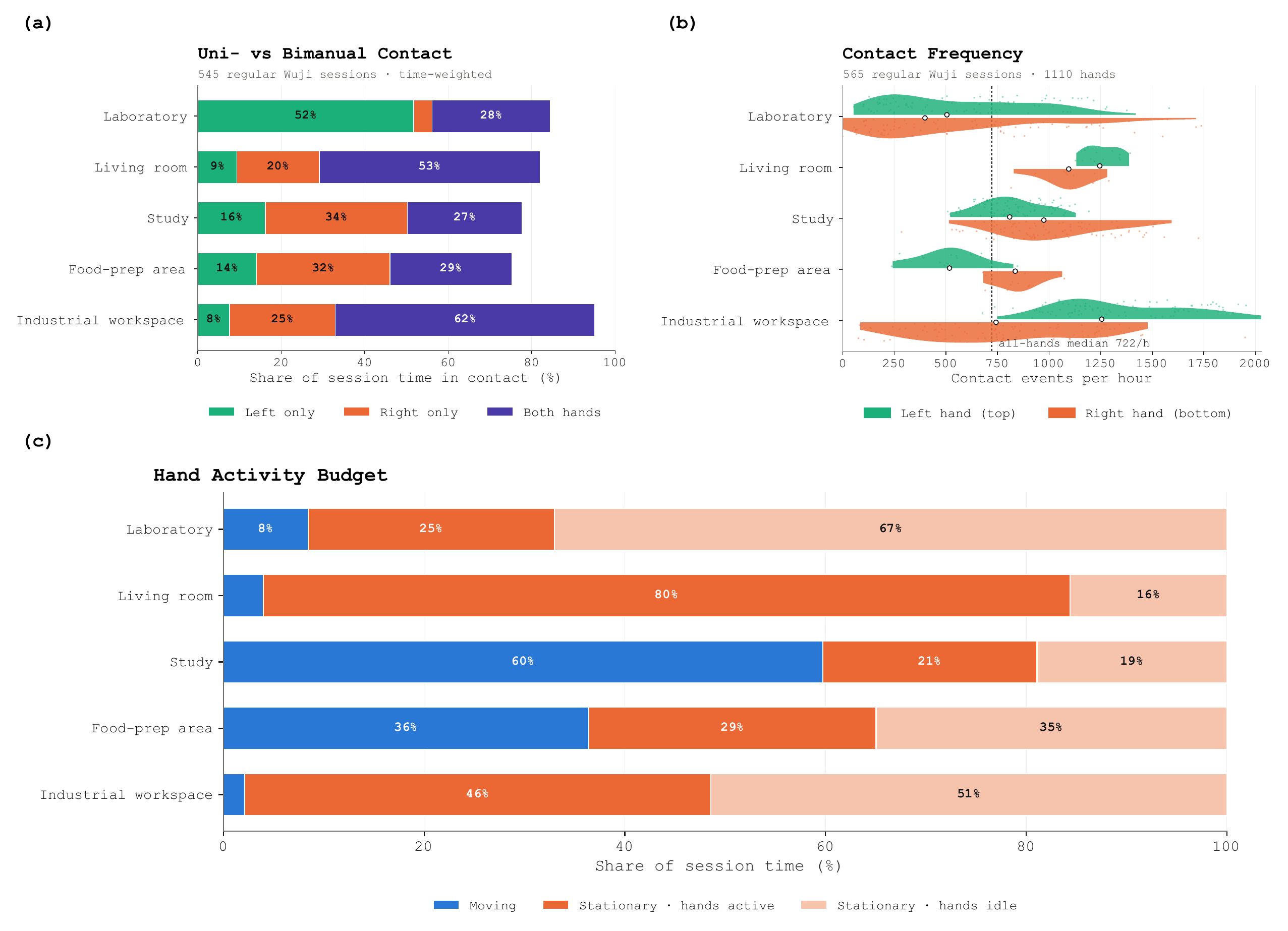}
\caption{Hand activity and contact: (a) unilateral/bimanual occupancy;
(b) event frequency; (c) locomotion and stationary hand activity.
Panels (a,b) retain healthy tactile streams.}
\label{fig:contact-statistics}
\par
\end{figure}

\textbf{Hand activity and contact.}
\cref{fig:contact-statistics} relates contact events and hand activity to
locomotion and stationary manipulation. Panel~(a) reports time-weighted
unilateral and bimanual occupancy over 545 regular Wuji recordings with
healthy tactile streams on both hands. Panel~(b) uses 565 recordings,
comprising 1{,}110 healthy hand streams; the median event frequency across
hands is 722 events per hour. Events last at least 0.1\,s, with gaps
shorter than 0.15\,s merged. Panel~(c) distinguishes locomotion from
stationary periods with active or idle hands, marking hands active above
150\,mm/s mean wrist speed. These summaries capture differences in task
structure: bimanual contact occupies 62\% of industrial recording time,
whereas laboratory recordings show more left-only contact (52\%).

\begin{figure}[!tp]
\centering
\includegraphics[width=0.95\linewidth]{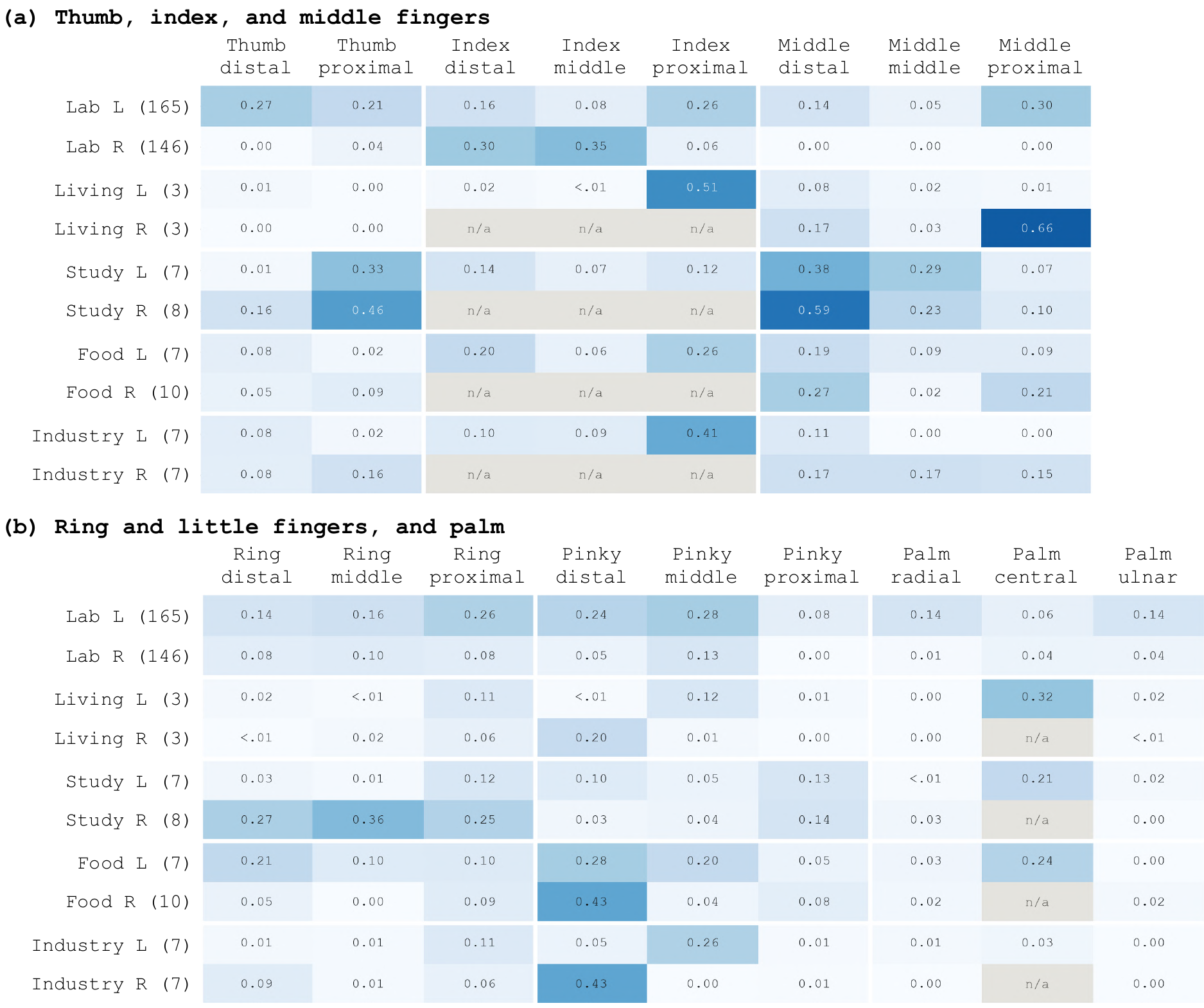}
\caption{Contact-frame occupancy by region: (a) thumb, index, and middle fingers;
(b) ring and little fingers, and palm. L/R denotes the hand; n/a indicates
insufficient live-taxel support.}
\label{fig:contact-occupancy}
\end{figure}

\textbf{Regional contact coverage.}
\cref{fig:contact-occupancy} resolves contact over the 17 anatomical regions
in \cref{fig:tactile-contact-regions}. Each value is the median active
fraction among contact frames. Multiple regions can be active together;
rows therefore need not sum to one. Parentheses give the number of analyzed
hands by environment and side. A region is n/a if fewer than three
recordings have at least 30\% live taxels there, including unsupported
right-hand index and central-palm regions.

\begin{figure}[!tp]
\centering
\includegraphics[width=0.99\linewidth]{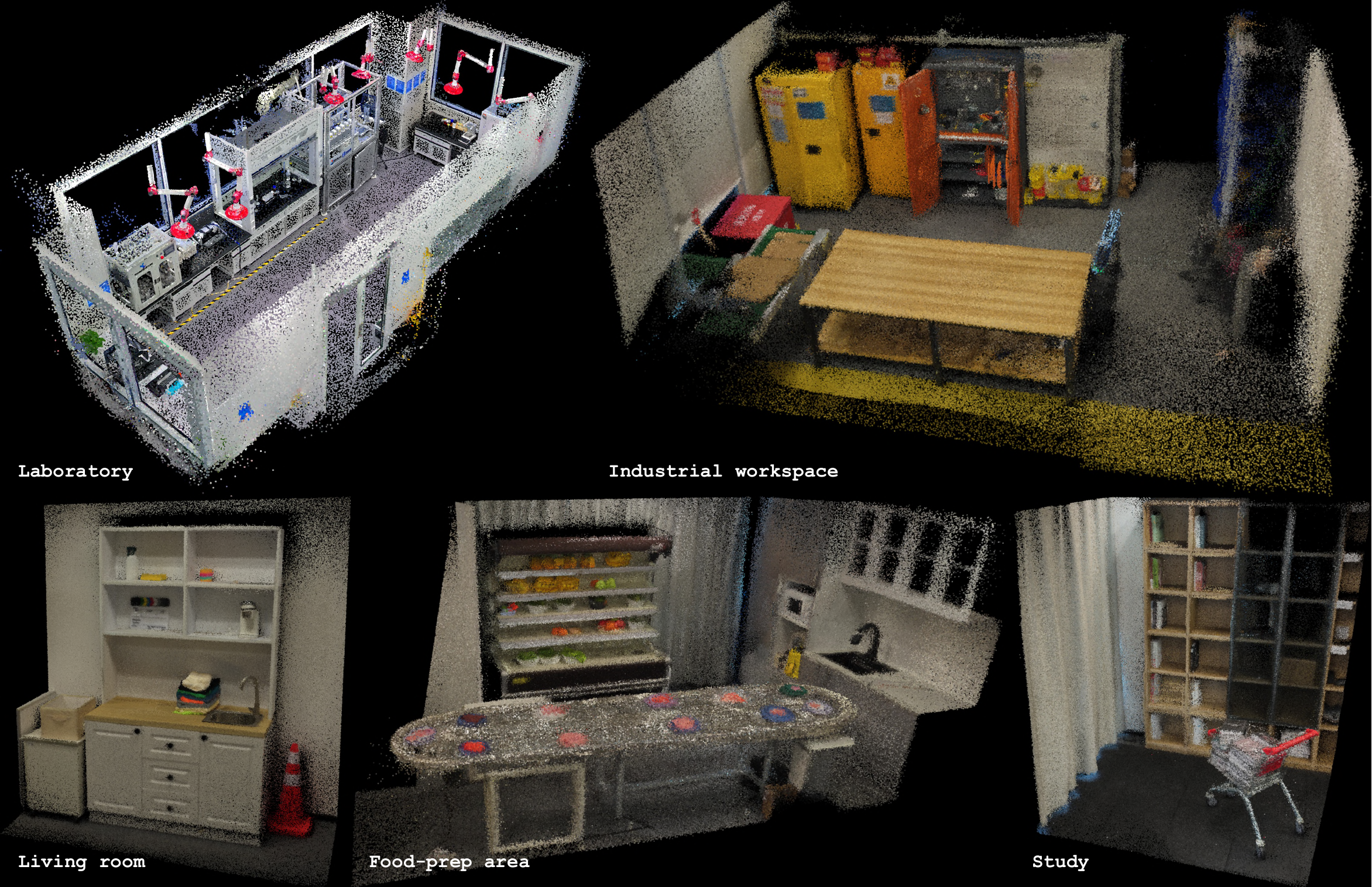}
\caption{Reconstructed capture environments: laboratory, industrial
workspace, living room, food-preparation area, and study. Each panel shows
one registered view.}
\label{fig:scenes}
\end{figure}

\textbf{Environmental context.}
\cref{fig:scenes} shows registered reconstructions of the five capture
environments. Together with the synchronized body--touch examples in
\cref{fig:scene-body-touch}, these views place the recorded motions and
interactions in their working spaces. Released geometry is limited by
capture quality and privacy requirements.

\textbf{Text coverage and vocabulary.}
\cref{tab:caption-statistics} describes the caption corpus. Temporal coverage
is the union of annotated intervals divided by total video duration, pooled
over recordings; it reaches 99.9998\%. Segment duration has median
7.486\,s and 90th percentile 24.000\,s. Captions average 98.87 characters,
with an exact duplicate rate of 0.0402\%, computed as
$(N-N_{\mathrm{unique}})/N$ over caption strings.
These measures describe coverage and granularity; semantic quality is
assessed in \cref{app:review-protocol,app:additional-experiments}.

\begin{table}[!tp]
\centering\footnotesize
\caption{Caption coverage, duration, length, and exact duplication.} 
\label{tab:caption-statistics}
\begin{tabular*}{\linewidth}{@{\extracolsep{\fill}}lr@{}}
\toprule
Statistic & Value\\
\midrule
Temporal coverage (video-duration weighted) & 99.9998\%\\
Segment duration p50 / p90 (s) & 7.486 / 24.000\\
Mean caption length (characters) & 98.87\\
Caption length p50 / p90 (characters) & 87 / 163\\
Exact duplicate rate & 0.0402\%\\
\bottomrule
\end{tabular*}
\end{table}

\begin{figure}[!htbp]
\centering
\includegraphics[width=0.68\linewidth]{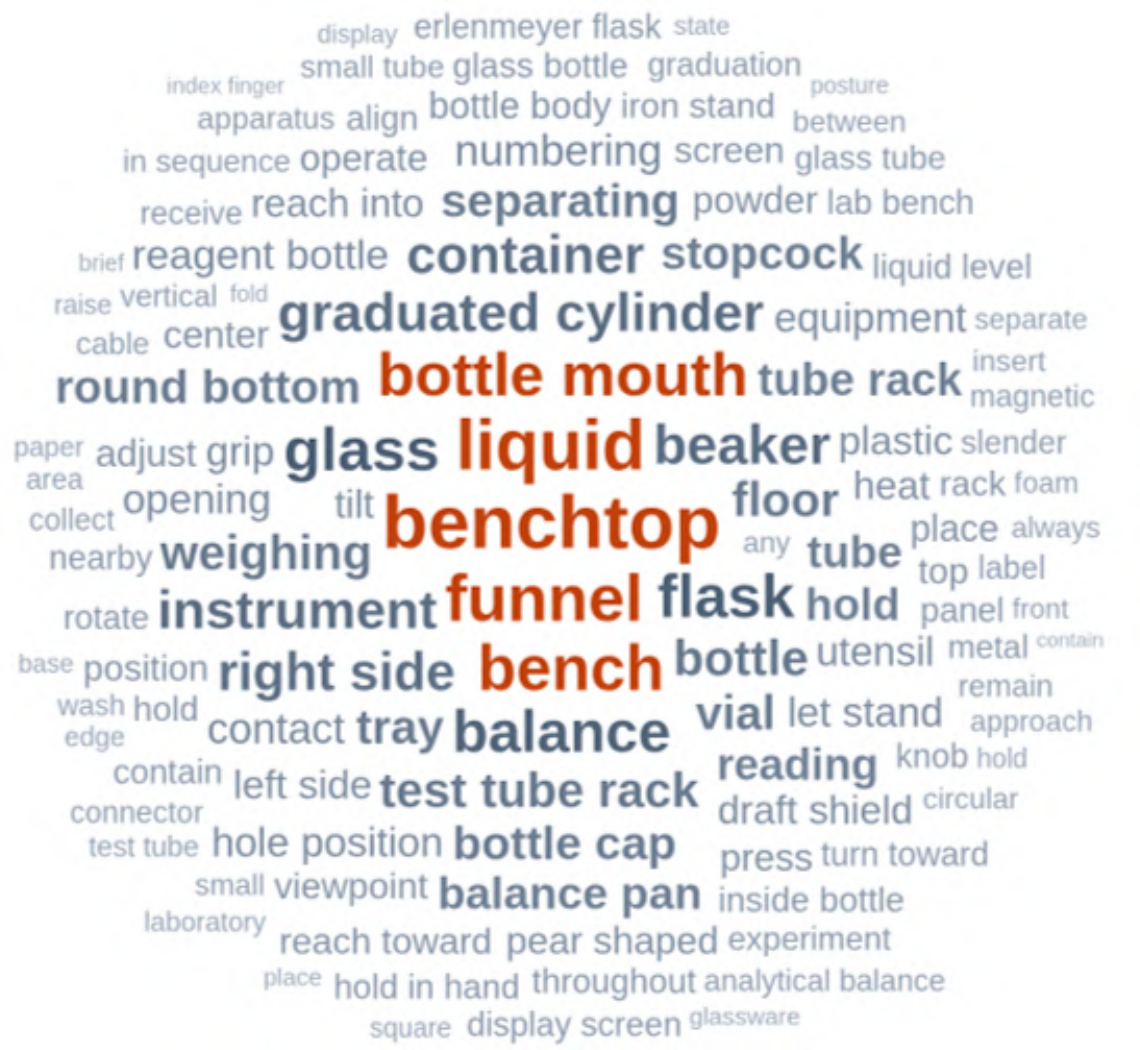}
\caption{Step-annotation vocabulary. Word size indicates term frequency;
frequent terms describe laboratory objects, equipment, and manipulation.}
\label{fig:task-anno-word-cloud}
\end{figure}

\cref{fig:task-anno-word-cloud} complements these statistics with the
frequency of terms in the step descriptions. Prominent words name
laboratory containers and equipment, including beakers, flasks, funnels,
and balances, together with actions such as weighing, holding, tilting,
and adjusting. Spatial terms and references to instrument readings add
context to these descriptions. This vocabulary reflects the strong
laboratory presence in the analyzed corpus; \cref{fig:text-annotation-visual}
shows how the descriptions are attached to individual task steps.

\FloatBarrier

\section{Experiments}
\label[appendix]{app:ablation}

\subsection{Evaluation setup}
\label[appendix]{app:evaluation-data}

\begin{figure}[!tp]
\centering
\includegraphics[width=0.99\linewidth,height=0.68\textheight,keepaspectratio]{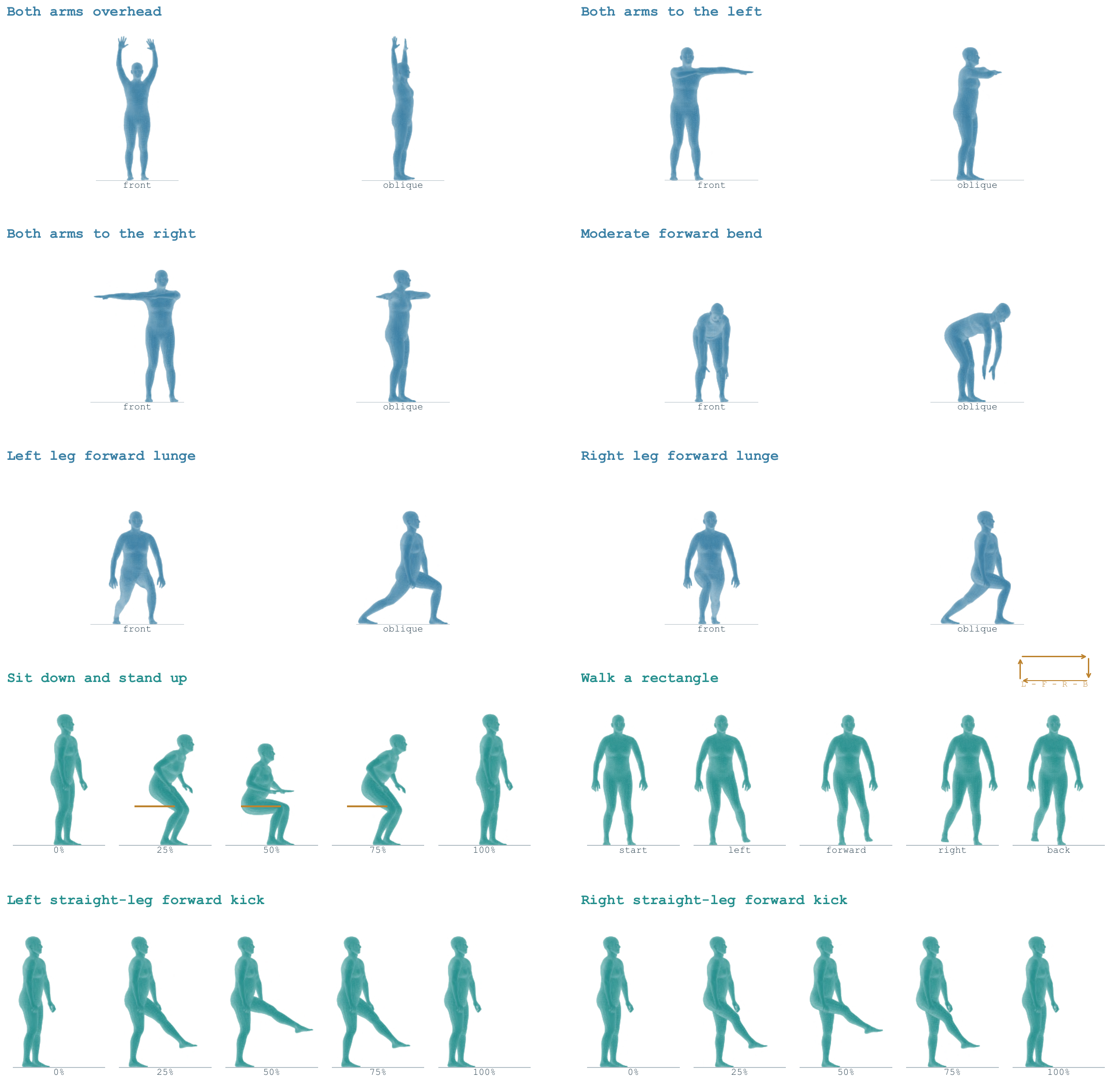}
\caption{Full-body action guides for the \evalsubset (B01--B10): static
holds followed by multi-phase actions.}
\label{fig:body-test-guides}
\end{figure}

The \evalsubset in \cref{sec:experiments} contains separate controlled
body, hand, and contact tests (\cref{tab:reference-actions}). Whole-body
capture covers six static and four dynamic actions
(\cref{fig:body-test-guides}), each repeated three times. Repetitions 1--2 support network adaptation; repetition 3 supplies
ten held-out test recordings. Hand tests cover ten right-hand configurations,
with three glove repetitions per configuration. The video comparison uses
30 recordings per method: 20 with a static camera and ten with camera
motion. These are separately recorded matched configurations, not
synchronized hand-pose references. Contact tests use button-marked press
intervals and task-specific anatomical regions. The detailed settings below
apply to the reported comparisons, rather than to the entire collection.

\begin{table}[H]
\centering\footnotesize
\caption{Action groups in the \evalsubset. Guides specify intended actions;
references and recording counts are defined separately for each modality.}
\label{tab:reference-actions}
\begin{tabularx}{\linewidth}{@{}llX@{}}
\toprule
Modality & IDs & Action coverage / controls\\
\midrule
Hands & H01--H03 & Spread, hook, fist; articulation and self-occlusion\\
Hands & H04--H07 & Thumb opposition to each of the other fingers\\
Hands & H08--H10 & Tripod pinch, lateral pinch, spherical grasp\\
Body, static & B01--B06 & Overhead, left/right cross-body reach, bend, left/right lunge\\
Body, dynamic & B07--B10 & Sit/stand, rectangular step sequence, separate left/right forward kicks\\
Contact & T01--T05 & Thumb--index press, bar press, spherical press, five-finger press, palm press; ON/OFF controls\\
\bottomrule
\end{tabularx}
\end{table}

The remaining \naturaldata have no matched accuracy references. Body
ablations use a 200-recording development pool, of which eight support
surface-term selection and the other 192 form the reported natural-data
cohort. This separation concerns constraint selection; it is not an unseen-task
or participant-generalization split. The protocol guides illustrate intended
actions; fitted body references and measured hand/contact references remain
separate from those illustrations.

\subsection{Whole-body reconstruction: inputs and references}
\label[appendix]{app:body-reference}

\textbf{Inputs and baseline adaptations.}
\cref{tab:baseline-contract} summarizes the inputs and controls of all
main-text comparisons. Public three-anchor baselines observe the head and hands after calibrated
device-to-joint conversion; rows marked as raw tracker poses omit that conversion.
Our system and the five-anchor AvatarPoser$^*$ and HMD-Poser$^*$ variants
also observe both ankles. AvatarPoser$^*$ uses a held-out checkpoint fine-tuned on repetitions~1--2 and
calibration recordings, and is evaluated before geometric refinement.
HMD-Poser~\citep{dai2024hmdposer} uses a local five-tracker adaptation trained on synthetic data without
official pretrained weights. SparsePoser~\citep{ponton2023sparseposer} additionally receives a pelvis pose
borrowed from our reconstruction. Comparisons use common body points and
timestamps; different sensing inputs are retained as explicit system differences.
The construction ablations instead fix the five-anchor inputs.

\begin{table}[H]
\centering\footnotesize
\caption{Inputs and reference controls for the main-text comparisons.}
\label{tab:baseline-contract}
\begin{tabularx}{\linewidth}{@{}>{\raggedright\arraybackslash}p{0.27\linewidth}>{\raggedright\arraybackslash}p{0.22\linewidth}>{\raggedright\arraybackslash}X@{}}
\toprule
Comparison & Inputs & Controls / reference\\
\midrule
Public body baselines & Head and wrists & Calibrated anchor conversion; raw-pose variants omit it\\
AvatarPoser$^*$ / HMD-Poser$^*$ & Five trackers & Local training; distinct training data and adapters (\cref{app:body-reference})\\
SparsePoser$^\dagger$ & Five trackers + borrowed pelvis & Assisted input; pelvis from our reconstruction\\
\name body reconstruction & Five trackers & Fixed profile; gloves are fused only after body fitting\\
HaWoR / HandFlow / glove & Video or glove skeleton & Matched held 
configurations; shared retargeter
(\cref{app:glove-evaluation})\\
HACO / pose rule / glove & RGB, hand pose, or touch & Same scored frames and regions; exclusions disclosed (\cref{app:glove-evaluation})\\
Text and skill review & RGB + instruction (+ skills) & Expert audit (\cref{app:review-protocol}); automated skill ablation (\cref{tab:skills})\\
Cosmos3-Edge & Joint generation & Preliminary training and conditioning studies (\cref{app:future-prediction})\\
\bottomrule
\end{tabularx}
\end{table}

\begin{figure}[!tp]
\centering
\includegraphics[width=0.99\linewidth]{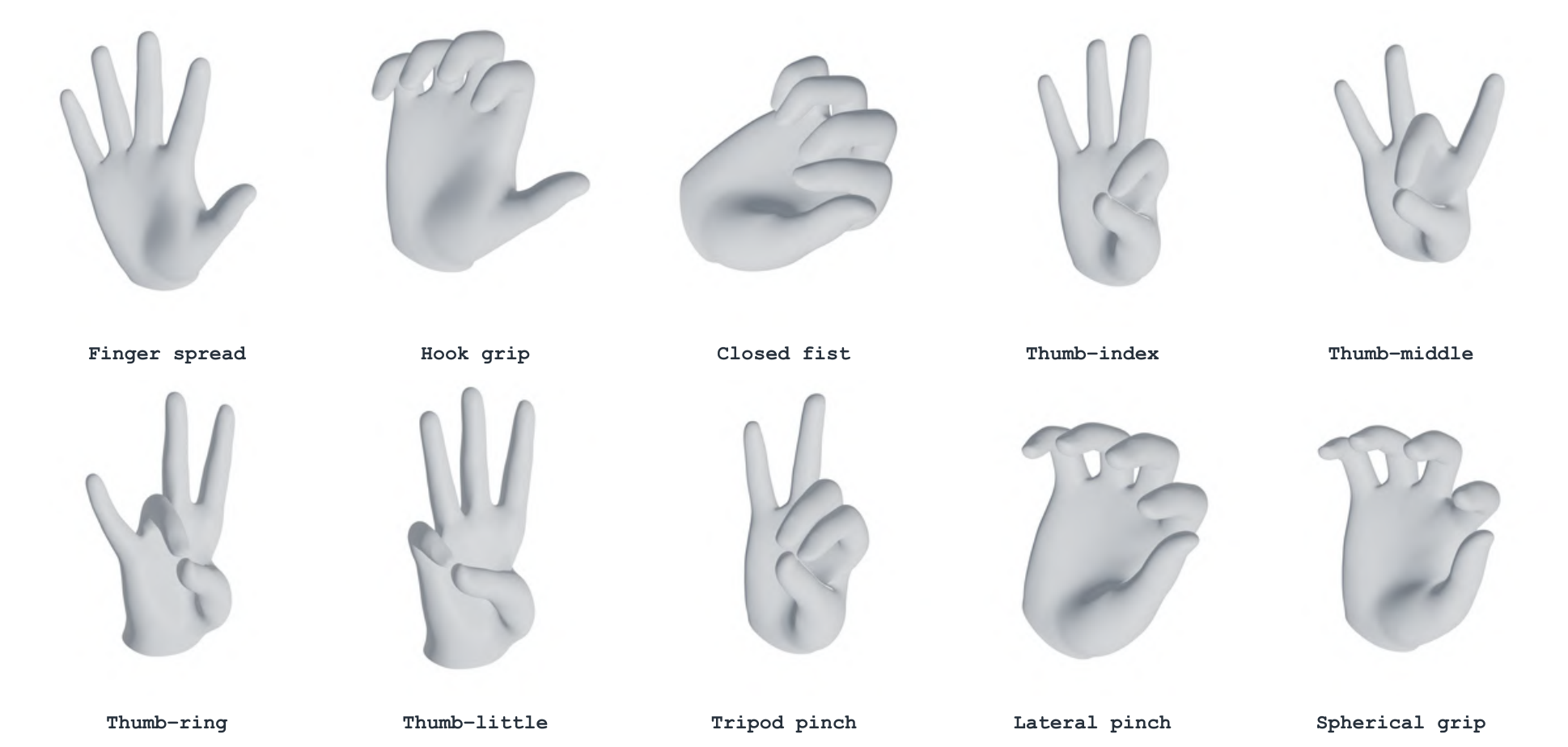}
\caption{Hand action guides for the \evalsubset (H01--H10, in order).
Generated MANO poses illustrate the instructions and local hand configurations.}
\label{fig:hand-test-guides}
\medskip
\includegraphics[width=0.99\linewidth]{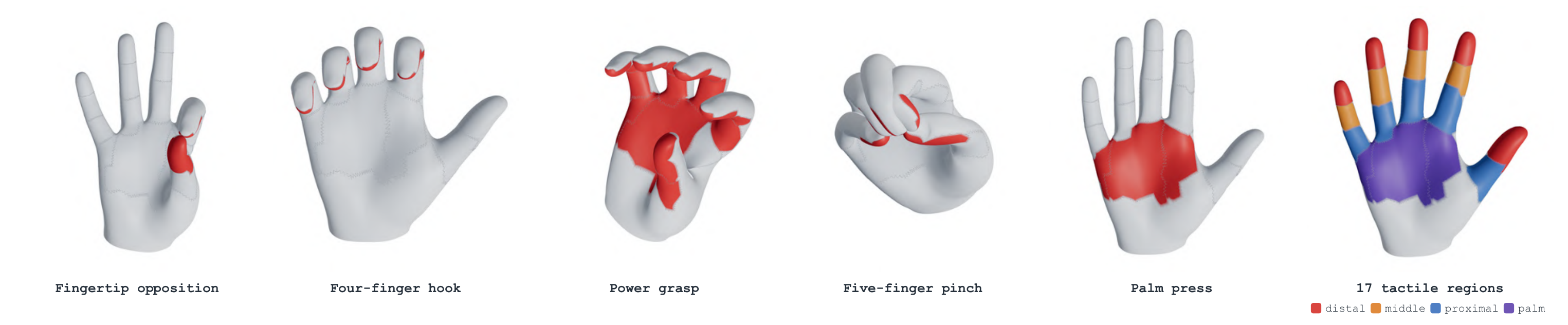}
\caption{Contact-action guides (T01--T05) and the 17-region anatomical map.
Red marks protocol-specified regions, not measured contact. Scoring uses
button-marked press intervals and the masks in \cref{app:glove-evaluation}.}
\label{fig:tactile-contact-regions}
\end{figure}

\textbf{Full-body reference labels for the \evalsubset.}
The controlled full-body recordings of the evaluation participant (ten actions
$\times$ three repetitions, \cref{tab:reference-actions}) do not have an independent optical reference. The accuracy numbers in \cref{tab:recon,tab:ablation} are therefore measured against \emph{reference labels}:
the same five-tracker refinement run for 300 iterations (learning
rate $0.01$, position weight $10^4$) with the participant's calibration, plus a
per-frame pose prior from the protocol library (weight 30 on hip, spine, knee,
ankle and foot rotations) that selects the intended articulation where the
tracker constraints alone admit several branches (forward bend, lunge, sit/stand, leg raise).
These labels agree with the trackers to within 5\,mm and were reviewed
frame-by-frame; the scoring uses a common body model and coordinate frame. The labels also
share our fitting model, calibration, and constraints, so agreement with them
is not independent ground-truth accuracy and can favor related methods. Baselines that do not observe the feet are still scored on the
same 22 body joints. Repetitions 1--2 train the participant-specific proposal
network; repetition 3 forms the full-body portion of the \evalsubset
(\cref{fig:recon-matched}). For this participant the joint-limit block
additionally contains a signed knee-hyperextension term, a pelvis-roll term
(both weight 100) and a 90$^\circ$ torso-lean limit so that deep bends are
admissible; evaluation on the \naturaldata keeps the production configuration.

For visualization, one instant per recording is selected from the fitted
reference, using the last of the candidate instants for peak deviation from
the initial stance, deepest knee flexion, strongest trunk lean, and lowest
pelvis position. All methods are shown at this same instant in
\cref{app:reconstruction-experiments}.

\subsection{Hand motion and contact capture}
\label[appendix]{app:glove-evaluation}

\textbf{Matched hand comparison.}
The configurations in \cref{fig:hand-test-guides} are captured with the
articulation glove and a RealSense D435 RGB stream ($1280\times720$, 30\,fps), reconstructed by
HaWoR and HandFlow \citep{zhang2025hawor,xu2026handflow}. All outputs use
the same retargeter, filter, and 20-DoF Wuji hand model. Each recording's
held configuration is the coordinatewise median of finite wrist-local
landmarks in its central hold interval. Ten glove-derived median robot
configurations define the task prototypes. Glove trials are scored with
the tested repetition excluded from prototype construction. Thus the
comparison tests pose retention and kinematic target recovery, without a
robot dynamics rollout.

\textbf{Contact references and exclusions.}
Button-held intervals mark perceived contact, and the T01--T05 instructions
specify the expected regions (\cref{fig:tactile-contact-regions}). Scoring
includes between-press negatives, excludes frames within $\pm100$\,ms of
button transitions, and excludes initial presses without a preceding rest
window. The same task-specific region and time masks apply to all methods;
regions outside the task mask are not counted as negative examples.

\textbf{Glove detector and comparison methods.}
For the controlled study, 120\,Hz readings are averaged into 30\,Hz video
bins. A taxel's baseline is its median during the 0.1--0.6\,s preceding a
press; its activation threshold is the maximum of 0.02, three robust rest
standard deviations, and the 99.5th percentile of rest fluctuations.
The health mask, estimated from earlier recordings, retains 526 of 744
taxels. A region is active when at least two and at least 5\% of its valid
taxels exceed their thresholds. HACO \citep{jung2025haco} vertex predictions
use the same count/fraction rule; a dark-glove image mask supplies the hand
box when its bare-hand detector fails. The pose-only rule uses the glove
skeleton: fingers must flex at least $40^\circ$ relative to the most open
pose and remain below 15\,mm/s over a 0.25\,s window for at least 0.2\,s.
Three or more flexed fingers indicate a grasp; otherwise only flexed
fingertips are marked. The glove detector uses annotated rest windows,
which should be considered when interpreting the system comparison.

\subsection{Evaluation metrics}
\label[appendix]{app:evaluation-metrics}

\textbf{Body accuracy (\cref{tab:recon,tab:ablation}).}
Body-point error averages Euclidean distances over 22 SMPL-X joints
(pelvis through wrists), in mm. Non-tracked error uses the 17 joints outside the head, wrists, and ankles;
the pelvis is included, with its error also reported separately as Root.
Root error is pelvis displacement in the fixed calibrated frame.
PA-MPJPE centers each pose, matches its Frobenius norm to the reference,
and applies the SVD rotation with reflection
correction~\citep{umeyama1991least}; the norm-matching
scale is not separately optimized by least squares. Velocity error averages
Euclidean differences of joint velocities, computed from adjacent positions
divided by their timestamp intervals, in mm/s. Scores use matched timelines
and finite observations, averaged within each recording and then equally
across recordings. No per-frame alignment is applied except for PA-MPJPE.

\textbf{Hand accuracy, stability, and continuity (\cref{tab:hand-quality}).}
Tip averages the Euclidean discrepancy of the five held fingertips in
wrist-local coordinates after matching hand length to the reference.
Hold RMS (root mean square) measures within-hold scatter: subtract each
fingertip's median held position from its emitted positions, square the
Euclidean distances, average across fingertips and valid hold frames, and
take the square root (mm). It is a stability measure, not independent pose
accuracy. Missing is the fraction of frames with no finite pose; Gap p95
is the 95th-percentile duration of contiguous missing runs pooled across
recordings. Success assigns the retargeted median pose to the closest of
ten glove-derived prototypes, using RMS landmark distance with weights
3 for tips, 0 for wrist, and 1 elsewhere. A trial succeeds when the nearest
prototype matches its instructed configuration; chance is 10\%.

\textbf{Sensor agreement and geometry (\cref{tab:ablation}).}
Tracker RMS squares Euclidean distances between predicted and observed
device positions, averages over valid device--frame pairs, and takes the
square root (mm); recording-level RMS values are then averaged equally.
Predictions use each variant's mount model, so this measures sensor
agreement. Crossing counts intersecting triangle pairs between distinct,
non-adjacent body parts using the same independent checker and 5\,Hz sample
times for all variants; frame counts are averaged within recordings, then
across recordings. Floor is the fraction of frames with any posed vertex
more than 5\,mm below the shared ground plane.

\textbf{Temporal quality (\cref{tab:ablation}).}
Slide is mean horizontal foot speed during common support intervals,
reported in mm/s. Support is inferred from raw ankles with horizontal speed
below 100\,mm/s, height within 30\,mm of that ankle's fifth-percentile
height, and duration at least 0.3\,s; pairs separated by more than 0.1\,s
are excluded. Jerk uses pelvis-relative joint positions in meters while
preserving world orientation. After resampling to 30\,Hz without bridging
invalid samples or gaps above 0.1\,s, take the third forward difference,
multiply by $30^3$, and average its vector norm over joint--frame pairs,
then equally over recordings (m/s$^3$). No smoothing is applied. Avg. is
the unweighted arithmetic mean of the nine displayed values in
\cref{tab:ablation}, including jerk and both data cohorts. This summary
depends on displayed units and has no single physical unit.

\textbf{Regional contact (\cref{tab:tactile}).}
After the shared masks in \cref{app:glove-evaluation}, true positives (TP),
false positives (FP), and false negatives (FN) are pooled over retained
frame--region decisions, including non-contact time. IoU is
$TP/(TP+FP+FN)$ and F1 is $2TP/(2TP+FP+FN)$; task-wise scores are not
averaged. Missing sensing support is excluded, not relabeled as no contact.

\subsection{Expert review of text annotations}
\label[appendix]{app:review-protocol}

\textbf{Language review.}
\cref{tab:language} reports expert review of model-generated text annotations prior to manual correction.
The audit covers 487 segments. Factual errors concern incorrect actions, objects, tools, or outcomes; unsupported claims lack evidence in the recording; missed and extra steps denote omissions and spurious additions, respectively; unjudgeable cases are reported separately.
Boundary error is the fraction of reviewed segments requiring temporal correction: 15 of 487 segments (3.08\%), reported at the segment level rather than as absolute timing deviation, with no frame-accurate boundary ground truth available.
Unjudgeable segments are those for which the available recording does not support a judgment.
Factual-error, unsupported-claim, extra-step, boundary, and unjudgeable rates
use reviewed segments; missed-step rate uses the reference steps that should
be described. These denominators need not coincide. All rates reflect pre-correction audit findings; human verification and correction follow model annotation.
The automated professional-skill ablation is reported separately in \cref{tab:skills}.

\subsection{Preliminary training study}
\label[appendix]{app:future-prediction}

\textbf{Pilot representation and metrics.}
\cref{tab:downstream} summarizes two preliminary Cosmos3-Edge~\citep{nvidia2026cosmos3} studies:
separate versus joint output generation and conditional action generation.
For the output comparison, video-only, action-only, touch-only,
and joint models all receive video/action/touch history. We report their
5,000-step checkpoints using the same sampled-evaluation implementation.
The separate-model row combines the video, action, and touch scores of three
independently trained models, one per output; it is not an ensemble prediction
or an average across models.
Action uses the
model's 57-dimensional hand-pose representation, derived from SMPL-X forward
kinematics: head and wrist poses plus wrist-local fingertip positions.
Each action coordinate value $\upsilon$ is normalized as
$\operatorname{clip}(2(\upsilon-\upsilon_{1\%})/\max(\upsilon_{99\%}-\upsilon_{1\%},10^{-6})-1,-1,1)$,
where $\upsilon_{1\%}$ and $\upsilon_{99\%}$ are its 1st and 99th percentiles
computed from training sessions only. Action MSE is
therefore an error in normalized feature space, distinct from the body-point
and fingertip errors in millimeters used for annotation evaluation. Touch
uses a separate token stream with regional contact, intensity, and validity
information; missing regions are masked.

\textbf{Training settings.}
The 5K configuration uses batch size 64 and AdamW with momentum coefficients
$(0.9,0.95)$, learning rates $5\times10^{-5}$ for the unfrozen backbone and
$5\times10^{-4}$ for new modality heads. Backbone weight decay is 0.01;
new heads use no decay. A 200-step warmup precedes cosine decay, with
gradient-norm clipping at 1 and seed 0. Training predicts clean targets from
noisy inputs: video-latent and action MSE have weight 1, regional contact
MSE weight 3, and tactile-intensity MSE weight 1.5. Invalid contact regions
are masked. Separate models activate only their own output losses, while
the joint model activates all three streams.

\textbf{Temporal split and scoring.}
The loader splits at session level within each task: sorted session indices
congruent to five modulo ten form validation; all windows from a session
remain in the same partition. The pilot uses 16 observed and 16 generated
samples at 15\,Hz, distinct from the release grid. Pixel-space PSNR evaluation
uses 256 held-out windows, two samples per window, and 20 denoising steps.
Prediction and target latents are decoded with the same clean historical
context; PSNR scores the generated interval against the VAE-decoded target,
not directly against uncompressed camera frames. PSNR is $10\log_{10}(1/\mathrm{MSE}_{\mathrm{pixel}})$ in dB for
pixels in $[0,1]$, averaged over windows and samples. Action MSE averages the 57 normalized
features, excluding padding channels. Contact predictions threshold the
mean of sampled binary states above 0.5; F1 pools TP, FP, and FN over valid
region--time entries as in \cref{app:evaluation-metrics}. The current output comparison in \cref{tab:downstream} reports higher
video PSNR and contact F1 and lower action MSE for joint generation.

\textbf{Conditional action generation.}
In the second study of \cref{tab:downstream}, recorded video, or video and
touch together, from the same interval as the generated action are supplied
as conditions. The unconditioned row receives no such additional
observations; the video/action/touch history remains available to all rows.
Because these conditions are observed over the action interval, this test
measures cross-modal conditioning rather than prediction from history alone.
Action MSE is computed in the same normalized feature space as above.

\subsection{Calibration and layered reconstruction ablations}
\label[appendix]{app:reconstruction-ablation}

Within each cohort of \cref{tab:ablation}, layer variants share the same
five-anchor inputs, proposal checkpoint, calibrated profile, and 60-iteration
budget. Proposal only skips refinement; Layer 1, Layer 1+2, Layer 1+3, and
Full activate the corresponding terms of \cref{eq:body-objective} without
retraining. Mean body replaces participant-specific shape and scale while
retaining mounts and the reconstruction procedure; proposals are recomputed
under that geometry. No mount calibration uses identity rotations and zero
attachment translations in training, inference, and refinement, with a
correspondingly trained network. Uncalibrated at test retains the calibrated
network but omits attachment correction during inference and refinement.

The natural-data cohort retains its production proposals. The independently
fine-tuned five-anchor AvatarPoser row in \cref{tab:recon} has its own
training pool and unrefined checkpoint; comparing its 64.1\,mm with the
79.0\,mm proposal-only row does not isolate refinement. The within-cohort
layer rows provide that comparison. Geometric scores must be read alongside
pose and tracker errors: moving the body toward sensor locations can reduce
crossing or floor violations without improving articulation.

\subsection{Professional-skill ablation}
\label[appendix]{app:additional-experiments}

\cref{tab:skills} compares unrevised annotation drafts generated with and
without professional Skills documents (\cref{app:text-skills}).
The annotation model, input videos,
task instructions, and decoding settings are held fixed; no model retraining
is involved. An automated judge flags factual errors and unsupported claims
at the window level, while event omission is measured against model-provided
reference events. Adding Skills documents lowers these three rates by
6.52, 5.52, and 7.05 percentage points, respectively. These automated
assessments are distinct from the expert audit in \cref{tab:language};
generated annotations still undergo human verification and correction.

\begin{table}[!tp]
\centering\footnotesize
\caption{Professional-skill document ablation with the same annotation
model. Automated judge results on unrevised drafts (\%; lower is better).}
\label{tab:skills}
\begin{tabular*}{\linewidth}{@{\extracolsep{\fill}}lrr@{}}
\toprule
Metric & Without Skills & With Skills\\
\midrule
Windows flagged for factual errors & 76.07 & \textbf{69.55}\\
Windows flagged for unsupported claims & 71.78 & \textbf{66.26}\\
Missed model-reference events & 48.97 & \textbf{41.92}\\
\bottomrule
\end{tabular*}
\end{table}

\subsection{Humanoid retargeting}
\label[appendix]{app:retarget}

\begin{figure}[!tp]
\centering
\includegraphics[width=0.99\linewidth,height=0.80\textheight,keepaspectratio]{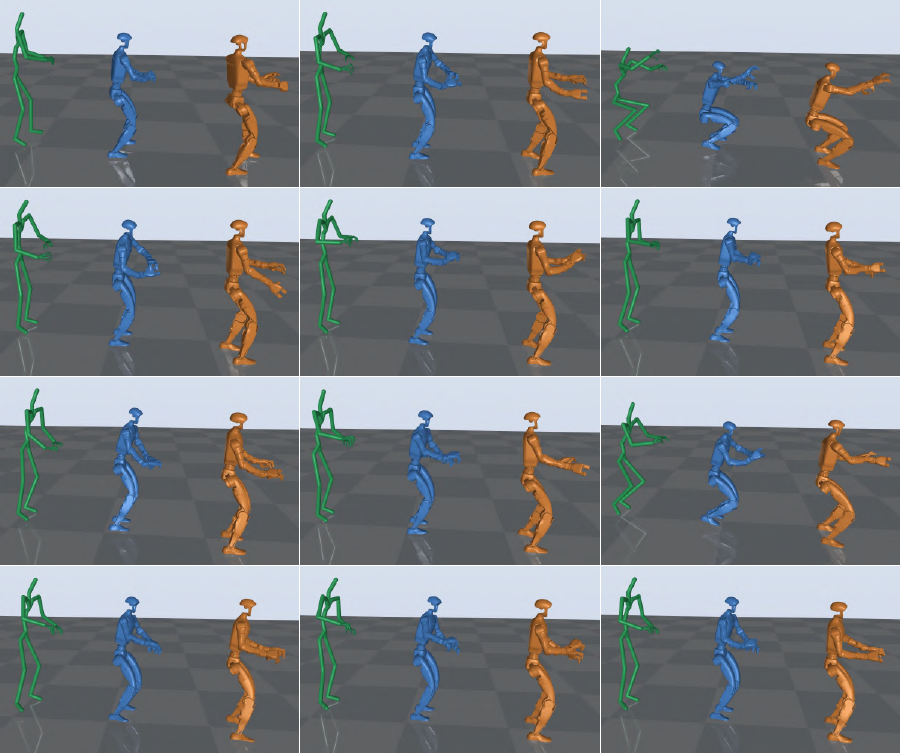}
\caption{Humanoid retargeting of a recorded episode. Green: reconstructed
human motion; blue: G1 kinematic retargeting; gold: physics-based whole-body
control in simulation. The trajectories are offset for visualization.}
\label{fig:retarget}
\end{figure}

Downstream humanoid control motivates preserving metric root translation,
per-joint rotations, and hand pose at a common sampling rate rather than
normalizing skeletons. \cref{fig:retarget} illustrates the transfer of a
135\,s reaction-setup recording to a Unitree G1, including reconstructed
SMPL-X motion, kinematic retargeting, and physics-based execution. The
kinematic stage uses GMR~\citep{ze2025gmr} whole-body inverse kinematics (IK) with position and
orientation targets, followed by root-height correction to prevent sole
penetration. We compare the corrected trajectories with GMR-scaled and
offset human targets at nine corresponding body origins: the pelvis, feet,
knees, elbows, and wrists. At temporally matched samples, the mean position
errors for the reaction-setup and reagent-weighing clips are 6.8 and
7.0\,cm, respectively; the corresponding wrist errors are 7.5 and 7.8\,cm.
These residuals include discrepancies between human joint centers and robot
link origins.
Position errors average Euclidean distances over the stated origins and
matched frames. The 14-body robot-reference score uses pelvis, torso, and
bilateral hip, knee, ankle, shoulder, elbow, and wrist link origins.
Root orientation is the angle of the relative quaternion, averaged over
frames; it is not a yaw-only error.

The physics-based stage follows the ZeroWBC~\citep{zerowbc2026} training
recipe and uses a reinforcement-learning tracking policy in MuJoCo~\citep{todorov2012mujoco}. Unlike
the original architecture, our implementation feeds short- and long-horizon
targets to an MLP and is trained only on the two clips. After 15{,}000 PPO~\citep{schulman2017ppo}
iterations, full-clip evaluation at 50\,Hz yields mean 14-body tracking
errors of 4.6 and 3.6\,cm after aligning torso horizontal position and yaw
with the robot reference, compared with 6.6 and 4.8\,cm without alignment.
No falls occur. Global wrist errors are 5.8 and 4.0\,cm. Root-position errors
are 3.8 and 2.5\,cm, and root-orientation errors are 1.8 and 1.3$^\circ$,
respectively. Relative to the same human targets and nine body origins used
above, the simulated trajectories have mean errors of 8.5 and 6.8\,cm and
wrist errors of 9.0 and 8.2\,cm, without evaluation-time alignment. Thus,
human-to-robot mapping error and robot-reference tracking error are measured
against different references and should not be added.

\textbf{Error sources.}
The post-IK floor correction raises the root by 2.7--6.7\,cm across the two
clips. Mean wrist error increases from approximately 5.6\,cm before
correction to 7.5 and 7.8\,cm afterward. Removing pelvis translation alone
still leaves wrist errors of 7.0 and 6.9\,cm and elbow errors of 9.8 and
9.5\,cm in the kinematic retargeting. Thus, global displacement does not
explain the full residual, and the current body mapping and IK trade-offs
leave substantial articulation error.
For the control policy, wrist and foot position rewards use
$\exp(-\overline{e^2}/0.3^2)$, with distances measured in meters. A 5\,cm
deviation therefore yields a reward of approximately 0.97. The joint-position
reward has weight 0.1, and no dedicated elbow- or knee-position rewards are
used. These choices provide limited incentives for centimeter-level
precision, and their individual effects have not been isolated by ablation.
The results therefore demonstrate preliminary simulated motion transfer
rather than precise object interaction. Joint optimization of ground contact
and body targets, together with tighter tracking rewards, remains an
important direction for future work.

\begin{figure}[!t]
\centering
\includegraphics[width=0.99\linewidth]{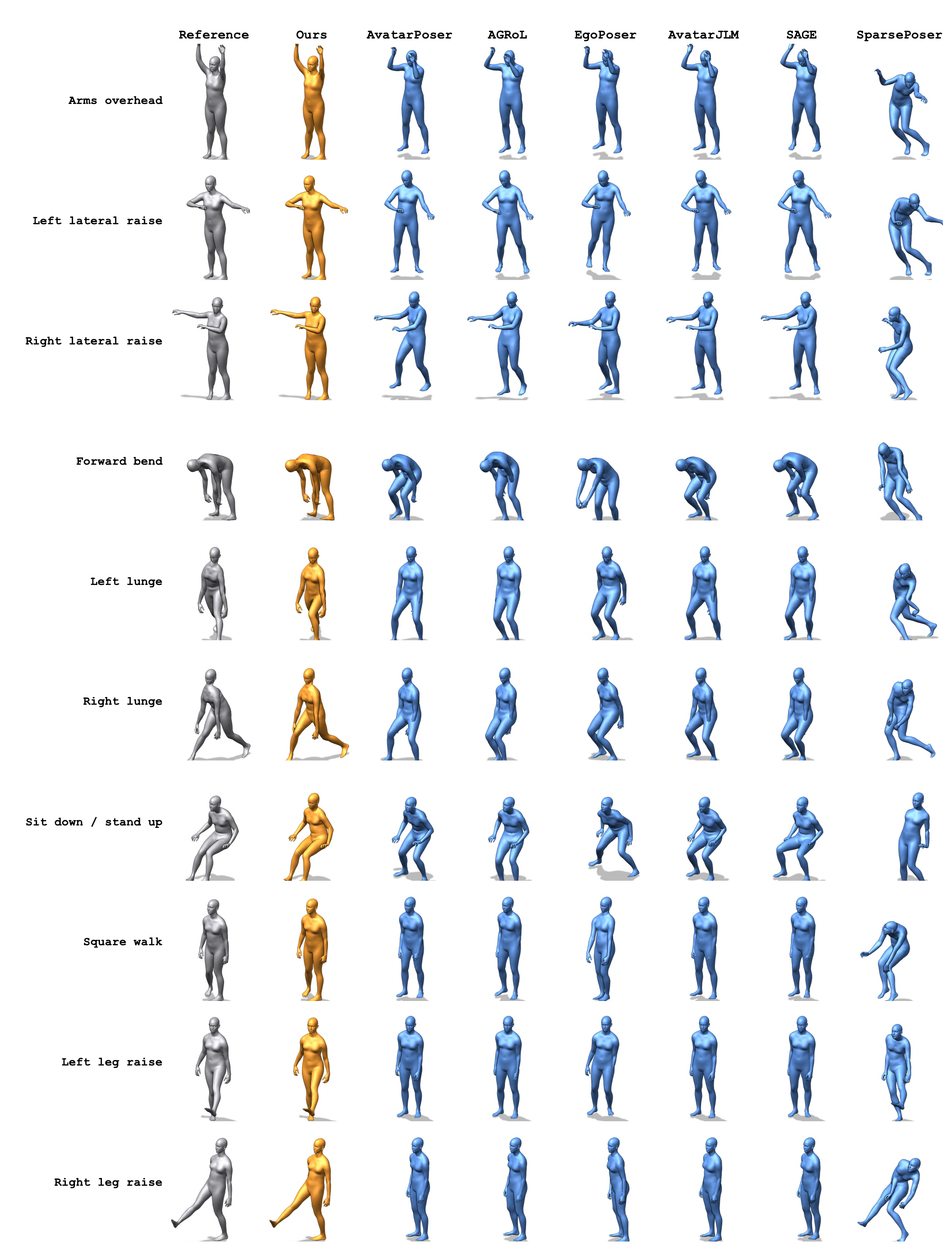}
\caption{Qualitative reconstruction comparison on the \evalsubset.
Each row shows the same instant for the fitted reference (grey), \name
(orange), and baselines (blue), rendered from a shared viewpoint.}
\label{fig:recon-matched}
\end{figure}

\textbf{Robustness, hands, and scope.}
A second rollout also remains fall-free under random base-velocity
perturbations applied every 2--4\,s, with horizontal components in
$[-500,500]$\,mm/s and yaw velocity in $[-0.5,0.5]$\,rad/s. Fingers are
retargeted independently to the three-finger hand by matching fingertip
vectors in the palm frame, yielding fingertip errors of 1.1--2.2\,cm. The
thumb--index aperture follows the human aperture with correlations of
0.991--0.993 (Pearson correlation of thumb--index fingertip distances
over matched frames). Hand poses are decoded using the same SMPL-X hand mean as in
glove fitting. Fingers are replayed kinematically, while the tracking policy
controls the 29 body joints.
Evaluation replays the training clips and therefore does not establish
generalization to unseen tasks. This example demonstrates transfer of
recorded motion, not hardware deployment; deployment would additionally
require state estimation and safety constraints, which are not evaluated
here.

\subsection{Qualitative examples}
\label[appendix]{app:reconstruction-experiments}

\cref{fig:recon-matched} compares \name with six baselines at matched instants
from the \evalsubset. Our results more closely recover the displayed forward
bends, lunges, and leg raises, including articulation between tracked endpoints.
Rows follow B01--B10 in \cref{tab:reference-actions}; the lateral raises,
square walk, and leg raises correspond to the cross-body reaches, rectangular
step sequence, and straight-leg forward kicks.
This visual comparison complements \cref{tab:recon}; the sensing inputs and
fitted references are specified in \cref{app:body-reference}.

\begin{figure}[!hp]
\centering
\includegraphics[width=0.99\linewidth]{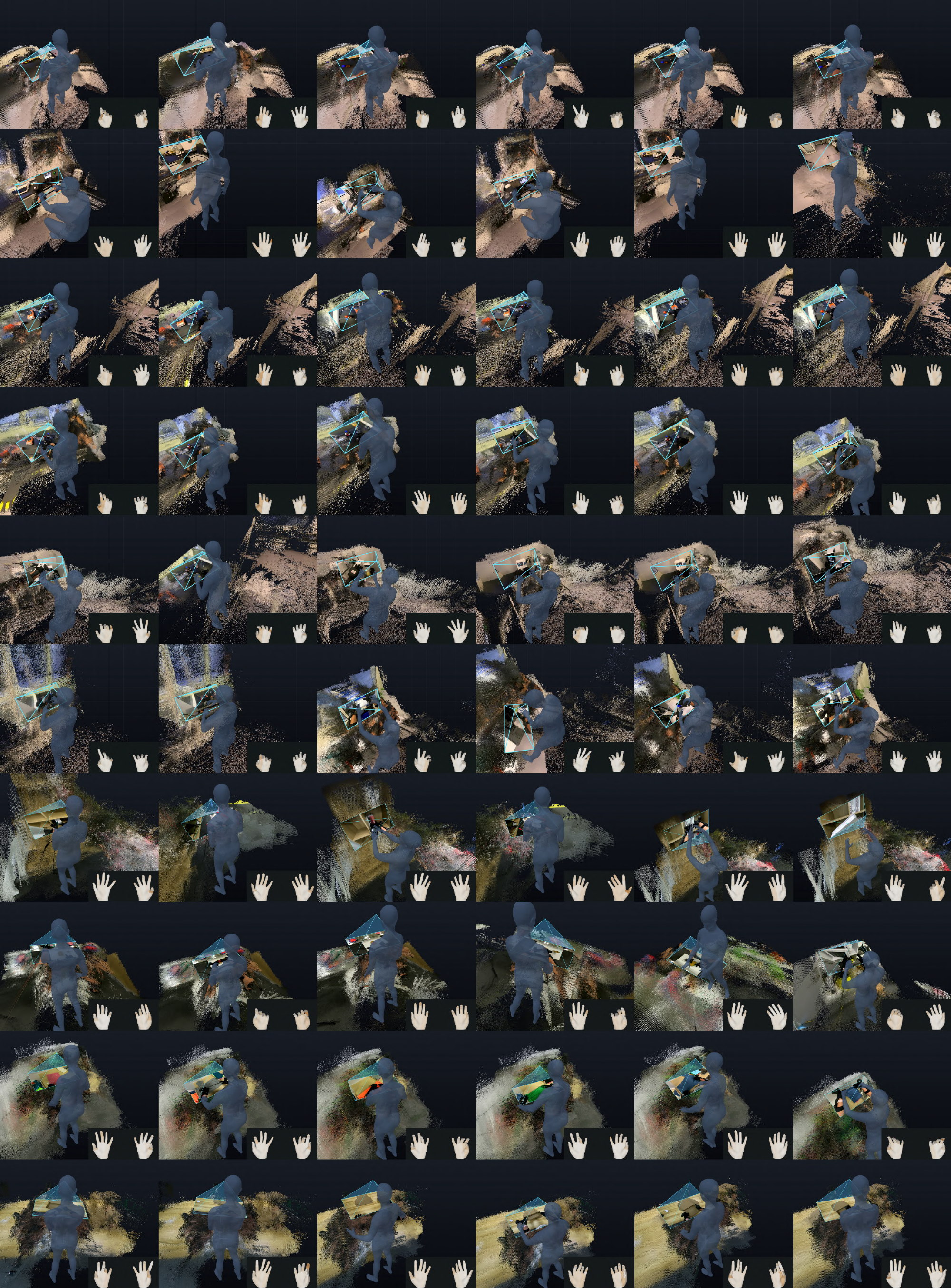}
\caption{Body and touch over time, ordered left to right. Rows:
reagent weighing, oven retrieval, reaction setup, extraction, rotary evaporation,
column chromatography, bookshelf organization, sandwich preparation, clothes
folding, and cardboard-box folding. Insets show both hands' synchronized touch;
gray-to-orange denotes baseline-relative increments.}
\label{fig:scene-body-touch-sequences}
\end{figure}

\textbf{Temporal body and touch examples.}
\cref{fig:scene-body-touch-sequences} expands the ten sessions in
\cref{fig:scene-body-touch}, with six synchronized instants per row,
evenly spaced from 10\% to 90\% of each recording. These sequences complement
the text studies with continuous body motion and contact observations.

\FloatBarrier

\section{Discussion on Capture Trade-offs}
\label[appendix]{app:cost-discussion}

\name combines portable deployment with accurate motion reconstruction and
complementary visual, body, hand, and tactile observations. Its wearable rig
supports collection in existing environments, while participant and mount
calibration grounds the recovered motion in measured geometry.
The body, hand, and contact comparisons in
\cref{tab:recon,tab:hand-quality,tab:tactile} support the quality of these
annotations relative to the evaluated baselines. Together, portability,
annotation quality, and sensing coverage motivate this design for capturing
human \locoman across tasks and workspaces.
\cref{tab:capture-deployment} summarizes the deployment trade-offs.

\begin{table}[H]
\centering\footnotesize
\caption{Capture arrangements and deployment requirements.}
\label{tab:capture-deployment}
\begin{tabularx}{\linewidth}{@{}lXX@{}}
\toprule
Approach / example & Sensing arrangement & Deployment requirement\\
\midrule
Multi-view RGB & External RGB cameras & Overlapping views, synchronization, camera calibration\\
HOT3D & Optical markers on hands and objects & Calibrated infrared-camera volume\\
Nymeria & Inertial body suit and Aria devices & Wearable sensors; no external camera array\\
ParaHome & IMU suit, gloves, and 70 RGB cameras & Instrumented camera volume\\
\name & Five inside-out trackers, ego camera, gloves & Body--sensor calibration; no external array or base stations\\
\bottomrule
\end{tabularx}
\end{table}

\textbf{Deployment and sensing.}
Multi-view RGB reconstruction provides an affordable route to whole-body
capture within a calibrated volume. FreeMoCap supports ordinary USB cameras
and recommends three to four views for general motion capture. Such arrays require overlapping views,
synchronization, and intrinsic/extrinsic calibration; moving the cameras
requires renewed calibration.
Multi-view methods can also recover hands, although occlusion and the small
image scale of fingers make fine articulation challenging
\citep{zhang2021lightweight}. Direct glove sensing complements body capture
and can additionally provide measured contact. Hybrid systems such as
ParaHome combine cameras, an inertial suit, and gloves
\citep{kim2024parahome}; HOT3D uses optical markers for hands and objects
\citep{banerjee2024hot3d}; Nymeria combines an inertial body suit with
wearable Aria devices \citep{ma2024nymeria}.
Our rig avoids external camera-array and base-station setup and reuses its
body--sensor calibration while mounts remain fixed. Its inside-out trackers
still depend on suitable visual conditions.

\textbf{Hardware costs.}
\cref{tab:capture-cost} reports costs in USD, excluding computers, software
licenses, mounting/cabling accessories, taxes, shipping, and setup labor; prices are public listings recorded in September 2026.

At USD~69.99 per C920s, four to six RGB cameras
cost less than the USD~1,184 tracker subsystem, making camera arrays attractive for fixed
capture spaces. The suit and optical-camera entries are hardware-only
listings. MANUS publishes no list
price, so its entry uses a reseller listing (September 16, 2026) excluding
the software license; the camera uses GoPro list
prices, USD~399.99 plus 79.99.
With the Wuji price converted to USD, subtotals
are approximately USD~7{,}979 (MANUS) and 5{,}397 (Wuji), trading added
wearable hardware for deployment flexibility and direct finger/contact
measurements.

\begin{table}[H]
\centering\footnotesize
\caption{Reference hardware costs in USD. Camera counts are illustrative;
glove options are alternatives.}
\label{tab:capture-cost}
\begin{tabularx}{\linewidth}{@{}Xlr@{}}
\toprule
Hardware / configuration & Quantity & Cost (USD)\\
\midrule
RGB array: Logitech C920s & 4--6 cameras & 280--420\\
\name body tracking: Ultimate Trackers + dongle & 5 + 1 & 1,184\\
Rokoko Smartsuit Pro II & 1 suit & 2,295\\
OptiTrack Flex 13 & 8 cameras & $\geq$8,792\\
\midrule
Ego camera: HERO13 Black + Media Mod & 1 set & 480\\
MANUS Metagloves Pro, non-haptic & 1 pair & 6,315\\
Wuji gloves & 1 pair & $\approx$3,733\\
\bottomrule
\end{tabularx}
\end{table}

\end{document}